\documentclass[sigplan,10pt]{acmart}

\usepackage{algorithm}
\usepackage{algpseudocode}
\algrenewcommand\algorithmicrequire{\textbf{Input:}}
\algrenewcommand\algorithmicensure{\textbf{Output:}}
\usepackage{multirow}
\usepackage{placeins}
\usepackage{xcolor}
\usepackage{colortbl}
\usepackage{tikz}
\usetikzlibrary{arrows.meta,positioning,calc,patterns,shapes.geometric,backgrounds,fit}
\definecolor{pcLayerPartition}{HTML}{6E7A82}
\definecolor{pcFixedParallel}{HTML}{86B0D6}
\definecolor{pcMutableParallel}{HTML}{1F5C99}
\definecolor{pcFixedReadyFirst}{HTML}{7FBE8E}
\definecolor{pcMutableReadyFirst}{HTML}{1E7B34}
\definecolor{pcFixedCostSelected}{HTML}{C795C2}
\definecolor{pcMutableCostSelected}{HTML}{8E3B8E}
\definecolor{pcFixedReactive}{HTML}{E59E63}
\definecolor{pcMutableReactive}{HTML}{C1500F}

\renewcommand\footnotetextcopyrightpermission[1]{}
\setcopyright{none}

\acmConference[EuroSys '27]{Proceedings of the European Conference on Computer Systems}{April 19--23, 2027}{Rabat, Morocco}

\begin{document}

\title{COPE: Composable Performance Emulation for Mixture-of-Experts Offloading Policies}

\author{\normalfont Hanming Zhang and Erhu Feng \\
{\normalsize\itshape Institute of Parallel and Distributed Systems, Shanghai Jiao Tong University} \\
} 
\renewcommand{\authors}{Hanming Zhang and Erhu Feng}
\renewcommand{\shortauthors}{Hanming Zhang and Erhu Feng}

\begin{abstract}

As LLMs scale, mixture-of-experts (MoE) architectures increase model capacity
while keeping per-token computation manageable through sparse expert
activation. This sparsity also enables memory-efficient deployment by retaining
most expert weights in CPU memory and transferring or executing only the routed
experts on demand. Existing MoE offloading systems typically target specific
model--platform configurations, yet differences in model sparsity, CPU and GPU
throughput, VRAM capacity, and host--device bandwidth can change the policy
ranking. Because implementing and exhaustively evaluating all candidate
policies incurs substantial engineering cost, efficiently selecting the
best-performing policy for each model--platform configuration remains a key
challenge for practical MoE offloading.

We present COPE
(\underline{C}omposable \underline{O}ffloading \underline{P}olicy \underline{E}mulator),
which selects the best-performing MoE offloading policy for a target
model--platform configuration before runtime integration. COPE comprises two
components. Its \textbf{Policy Matrix} factors the design space into two
independent axes, expert residency and MoE execution strategy, enabling
mechanism reuse across candidate policies and exposing four new candidates.
Its \textbf{Virtual Heterogeneous Machine} profiles combined-operator schedules
on the target platform and combines these measurements with offline routing
traces to estimate candidate performance. Compared with the evaluated
baselines, COPE improves fastest-policy selection accuracy by 24\%--100\% and
reduces performance-prediction error by $1.7\times$--$6.8\times$.
By identifying the preferred policy before
integration, COPE further reduces policy-specific implementation effort by up to 74\%.

\end{abstract}

\maketitle
\pagestyle{plain}
\thispagestyle{plain}

\section{Introduction}

Sparse mixture-of-experts (MoE) architectures have become a widely adopted
approach to scaling LLM capacity without proportionally increasing per-token
computation. A router activates only a small subset of experts for each token,
allowing Switch Transformer~\cite{switch-transformers},
Mixtral~\cite{mixtral}, DeepSeek-V2~\cite{deepseek-v2}, and
Qwen3~\cite{qwen3} to combine a large parameter space with a much smaller
active working set.
This sparsity also creates a natural opportunity to tier expert weights across the memory hierarchy: CPU RAM can retain the complete expert set, while GPU VRAM holds only a selected subset.
Academic systems and production runtimes~\cite{fiddler,mixtral-offloading,moe-infinity,promoe,hybrimoe,dali,freetoken} have consequently developed a range of expert offloading, caching, and prefetching mechanisms for deploying MoE models whose weights exceed GPU memory capacity.

However, deploying MoE models through offloading is considerably more complicated in practice.
This complexity arises from three sources: variations in model sparsity, the diversity of offloading mechanisms, and the heterogeneity of hardware configurations.
First, although MoE architectures have been widely adopted by recent LLMs, different models exhibit substantially different routing sparsity and expert-popularity distributions.
Consequently, the number of experts activated per token and the distinction between frequently and infrequently accessed experts vary across models.
Second, existing offloading systems adopt fundamentally different mechanisms. For example, llama.cpp~\cite{llama-cpp} performs sequential offloading at layer granularity, vLLM~\cite{vllm-uva} uses a unified CPU--GPU address space to access host-resident weights,
and KTransformers~\cite{ktransformers} assigns different categories of model weights to different processors.
Finally, hardware heterogeneity further complicates policy selection. MoE offloading often targets edge systems, including mobile phones, personal computers, and AI cubes. These platforms combine diverse CPU, GPU, and NPU architectures, providing no uniform hardware profile.

As a result of this complexity, \emph{\textbf{no single offloading strategy} performs best across \textbf{all model--platform configurations}}.
A strategy that performs well in one setting can exhibit substantially different behavior in another.
Implementing every candidate offloading strategy on each target platform would incur substantial engineering cost. Therefore, efficiently selecting the best-performing strategy for a given model--platform pair remains a key challenge for edge MoE deployment.

We present COPE (\underline{C}omposable \underline{O}ffloading \underline{P}olicy \underline{E}mulator), which structures the policy space and
selects the MoE offloading policy expected to perform best for a
target model--platform configuration before runtime integration.
COPE answers two questions: (1) How can offloading designs be organized into reusable choices,
and (2) which candidate is expected to deliver the best performance?
To answer the first question, COPE introduces the \textbf{Policy Matrix}, which decomposes an MoE offloading design into two composable dimensions: expert residency and MoE execution strategy.
Expert residency determines where each expert is stored in the memory hierarchy,
including static schemes such as layer-wise and hot/cold placement,
as well as dynamic schemes that update expert residency at runtime.
Expert execution strategy determines where and how each routed expert is computed,
including CPU execution, weight transfer followed by GPU execution, and concurrent CPU--GPU execution, etc.
The mechanisms in these two dimensions can be implemented independently and reused across policies.
Combining one choice from each dimension yields a concrete offloading policy.
This factorization exposes four candidate policies not covered by existing MoE offloading systems.

The performance ranking of candidate offloading policies is highly sensitive to both model characteristics and the target hardware platform. Differences in CPU and GPU performance, interconnect bandwidth, synchronization overhead, and model sparsity can change which policy performs best. Selecting the best policy for each platform by implementing and benchmarking every candidate on real hardware would incur prohibitive engineering costs.
Therefore, to answer the second question, COPE provides the \textbf{Virtual Heterogeneous Machine} (VHM).
VHM profiles atomic compute, data-transfer, synchronization, and asynchronous-dispatch operators on the target hardware, as well as combined-operator schedules that capture their interactions.
These measured costs absorb differences in accelerator speed, bandwidth, and overlap.
Given the profile and timing-free routing traces containing routed and predicted
expert IDs, VHM replays each candidate's
cache states and computes its predicted TPOT without consulting hardware specifications again.
Across hardware configurations, COPE reuses the same offloading-policy mechanism in VHM,
thereby reducing the engineering effort for new target platforms.

We evaluate COPE's policy-selection accuracy and its potential to reduce
engineering effort through two complementary implementations. First, we
implement nine Policy Matrix candidates, including those newly exposed by
COPE, in llama.cpp~\cite{llama-cpp}. We benchmark these implementations on
three CPU--GPU platforms (A100, RTX~4090, and RTX~5090) and across 11 additional controlled
A100 resource configurations. The end-to-end measurements serve as ground
truth for policy selection. Second, we implement the same nine
policies in the Virtual Heterogeneous Machine using existing computation-graph
and operator interfaces, and compare its TPOT predictions with the measured
ground truth.

COPE selects the measured minimum-TPOT candidate in 134/150 A100, 20/20
RTX~4090, and 42/45 RTX~5090 cases. Compared with the Hierarchical Roofline Model~\cite{moe-lightning},
COPE reduces TPOT-prediction MAPE across
the evaluated models from 61.53--68.52\% to 9.39--17.43\% on the RTX~4090,
and from 47.21--53.92\% to 7.95--13.34\% on the RTX~5090.
By identifying the best-performing candidate before runtime integration, COPE
avoids unnecessary policy-specific engineering effort.
For Qwen-3 on the RTX~4090, implementing only the COPE-selected
\emph{Fixed Parallel} policy reduces policy-specific code by 74\% compared
with implementing all candidates using the Policy Matrix's shared mechanisms.

\section{Background}
\label{sec:background}

\subsection{MoE Inference with Offloading}

An MoE layer routes each token to a top-$k$ subset of its experts. Sparse
activation reduces per-token arithmetic, but any expert may be selected by a
later token, so the complete parameter set must remain available. In a
partial-fit deployment, CPU RAM holds all model weights while GPU VRAM holds
attention, activations, the KV cache, workspace, and only some expert weights.
A router selection without a GPU copy becomes a cache miss, which incurs
host-side work or weight movement.

A cache miss exposes several exact-computation paths. Keeping a whole MoE
layer on the CPU avoids repeated weight transfers but runs even popular
experts there~\cite{llama-cpp}. With a partial GPU cache, a resident expert
runs on the GPU while a miss runs on the CPU and returns its result
\cite{fiddler}; alternatively, the selected weights are staged to the GPU,
trading host computation for weight movement~\cite{mixtral-offloading}.
Mapped host weights permit GPU reads without explicit staging, but those
reads still cross the interconnect~\cite{vllm-uva,cuda-memory}. All four
paths retain the router choice and stored precision.

Overlap changes, but does not eliminate, these costs. GPU hits may run while
CPU misses compute or weights transfer; asynchronous copies need pinned host
buffers, and actual overlap depends on hardware and synchronization
\cite{cuda-async}. Every selected result must join before the layer finishes.
Expert size, CPU/GPU throughput, link bandwidth, and cache hits therefore
jointly determine the fastest path.

This trade-off is acute in local batch-1 decode. Sparse activation permits
models whose total weights exceed VRAM, while local execution keeps prompts
and outputs away from remote services. General-purpose serving runtimes such
as vLLM, SGLang, and TensorRT-LLM optimize broader serving efficiency
\cite{vllm-sosp,sglang-paper,tensorrt-llm}, not this particular cache-miss
decision. With one token in flight, CPU work, weight movement, GPU work, and
coordination remain visible in time per output token (TPOT).

We therefore study batch-1 decode when the model fits in host memory but not
entirely in VRAM. We say a
\emph{model--platform} fixes the model, device, and runtime; a
\emph{configuration} is a resource slice of one platform.

\subsection{Existing Offloading Policies}

\begin{figure}[htp]
  \centering
  \setlength{\belowcaptionskip}{-5pt}
  \includegraphics[width=\columnwidth]{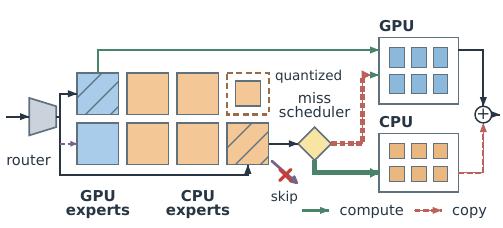}
  \caption{Expert offloading choices. RAM retains all experts; VRAM caches some
  copies. Solid paths preserve the checkpoint: a hit runs on the GPU, while a
  miss runs on the CPU or stages to GPU. Dashed rerouting, a crossed-out skip
  arrow,
  and a smaller quantized CPU weight illustrate alternatives outside our scope.}

  \label{fig:expert-cache-anatomy}
\end{figure}

Existing MoE offloaders accelerate expert computation by changing where experts
run or by modifying what the checkpoint computes. In the first case, a routed
expert can run on the CPU or have its weights staged for GPU execution
\cite{fiddler,mixtral-offloading}. Other approaches skip selected experts
\cite{not-all-experts-equal}, substitute a cached expert for the router's choice
\cite{smoe}, or quantize expert weights~\cite{moe-apex}; these changes may alter
the checkpoint's outputs. We study \emph{exact-routing} offloading: the
checkpoint's router decides which experts run, every selected expert executes,
and its stored precision is retained. Figure~\ref{fig:expert-cache-anatomy}
contrasts these choices.

Expert placement and prefetching can further reduce cache misses without
changing the model's computation. Fiddler uses expert popularity measured on a
calibration workload to keep frequently selected experts in VRAM~\cite{fiddler}.
Lookahead offers another opportunity: DejaVu observes that representations
change slowly across neighboring Transformer layers~\cite{deja-vu}, while
ProMoE predicts future MoE expert use from intermediate states and prefetches
their weights~\cite{promoe}. Such predictions change which weights are cached,
not which experts the checkpoint router eventually selects. We therefore also
study the effects of placement and prefetching on batch-1 decode.

\subsection{A Growing Space of Models and Configurations}

Target platforms span consumer GPUs such as RTX~3090, RTX~4090, and RTX~5090,
workstation accelerators such as L40S, and other vendors' devices. GPU throughput, VRAM, and
host--device bandwidth do not scale together; neither do CPU core count and
memory bandwidth. Even one platform admits different CPU-worker, GPU-share, and
VRAM-budget configurations.

Models vary independently along four policy-critical quantities. Mixtral uses
8 experts/top-2 and DeepSeek-V2-Lite 64/top-6. Qwen3-30B-A3B uses 128/top-8;
Qwen-3.5-35B-A3B uses 256/top-8 \cite{mixtral,deepseek-v2,qwen3,qwen35-35b-a3b-card}. The latter
three have 26, 48, and 40 MoE layers. F16 split gate/up/down experts occupy 16.5~MiB
in our DeepSeek-V2-Lite target and 6~MiB in Qwen-3.5, while Mixtral's geometry
implies about 0.18B parameters (336~MiB F16) per expert FFN. Total/active
parameters range from about 16B/2.4B for DeepSeek-V2-Lite to 46.7B/12.9B for
Mixtral \cite{mixtral,deepseek-v2}. These four quantities determine cache
capacity, miss work, transfer bytes, and how many layers repeat them.

These examples show breadth; every target must pass its model--device gates.
Policy rankings also cannot be extrapolated
between models or configurations. The real question is which feasible template
to integrate for a given model--platform without first implementing them all.

\subsection{LLM simulators.}
Existing simulators target broader deployment questions. Vidur explores
inference scheduling, batching, and tensor-parallel configurations
\cite{vidur}; LLMServingSim~2.0 and Frontier model heterogeneous or
disaggregated serving, including multi-GPU parallelism and prefill--decode or
attention--FFN separation \cite{llmservingsim2,frontier}. SimAI and
ASTRA-sim2.0 instead emphasize large-scale training, collective communication,
and GPU-interconnect modeling \cite{simai,astrasim2}.
Although Frontier and LLMServingSim~2.0 support MoE routing or expert
parallelism, their evaluations do not address our central question: how
different MoE offloading policies affect inference performance on
memory-constrained edge devices. In particular, neither simulator integrates
and compares the multiple MoE offloading policies in
Figure~\ref{fig:policy-space} within a common model. COPE targets this
narrower, policy-level prediction problem rather than general multi-GPU serving
or training simulation.

\section{Observations}
\label{sec:observations}

\subsection{Observation I: Model--Platform Dependent Policy Performance}

\textbf{The Winning Policy Changes with the Model and Platform.}
The best MoE offloading policy depends on which operations lie on the decode
critical path. CPU and GPU throughput, interconnect bandwidth, and VRAM capacity
determine the costs of computation, staging, and expert residency. Expert
size, the number of routed experts per token and routing locality
determine how frequently these costs are incurred. Changes in either the
resource configuration or the routing workload can therefore shift the
bottleneck and reverse the ordering of candidate policies.

\begin{figure}[t]
  \centering
  \setlength{\belowcaptionskip}{-10pt}
  \includegraphics[width=\columnwidth]{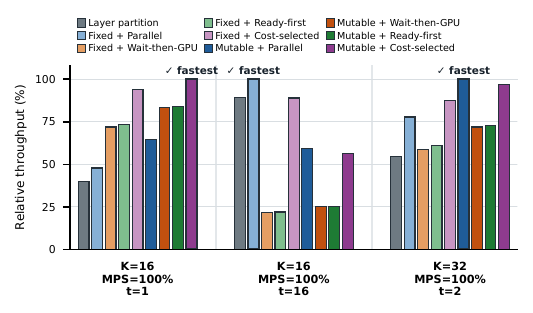}
  \caption{DeepSeek-Coder-V2-Lite policy landscape for three resource slices
  on one A100. Throughput is normalized to the fastest candidate in each slice;
  color identifies execution and hatching identifies residency.}

  \label{fig:policy-landscape}
\end{figure}

Figure~\ref{fig:policy-landscape} isolates resource sensitivity for
DeepSeek-Coder-V2-Lite on a single A100. With the model and physical GPU
unchanged, the measured winner shifts as the CPU resource configuration changes.
\emph{Mutable cost-selected} wins with one CPU worker, \emph{fixed Parallel} with 16 CPU workers,
and \emph{mutable Parallel} with two workers and a larger $K=32$ expert cache. This
comparison shows that resource availability alone can reorder the candidate
policies.

We further validate this configuration-dependent behavior across multiple
models and hardware platforms.
Sweeping CPU workers, requested MPS share, and VRAM capacity across three models produces multiple winners
(Figure~\ref{fig:hardware-tradeoffs} and
Table~\ref{tab:winner-grid}). Changing the physical platform produces a further
shift. Across the evaluated A100 configurations, candidates using per-expert GPU
residency achieve the lowest measured TPOT. At the evaluated RTX~4090
configuration, with a fixed CPU allocation and full GPU compute share,
\emph{layer partition} is fastest for three of the four models, while
\emph{Fixed Parallel} wins for Qwen3.

\textbf{Simple Policies Can Beat Complex Ones.}
The controlled A100 study evaluates whether more adaptive offloading mechanisms
consistently reduce TPOT. The results show that additional adaptivity does not
always translate into better performance.
\emph{Fixed Parallel}
achieves the lowest measured TPOT in both 16-worker configurations,
outperforming candidates that use mutable residency or cost-selected execution.
Even \emph{layer partition}, the simplest candidate, is only 12.2--25.0\%
slower than the winner (Section~\ref{sec:policy-variation-evaluation}).

These results show that additional policy mechanisms do not necessarily justify
their runtime overheads. Although COPE optimizes TPOT rather than engineering
complexity, simple designs must remain explicit candidates because they can be
competitive or even optimal under specific resource configurations.

\subsection{Observation II: Composable and Executable Policy Space Modeling}

\textbf{Policies Decouple into Two Finite Dimensions.}
Prior systems have explored a range of MoE offloading mechanisms, including
demand-driven expert loading, CPU--GPU co-execution, proactive caching, and
joint scheduling and cache management
\cite{mixtral-offloading,fiddler,promoe,hybrimoe,dali,freetoken}. Although these systems
adopt different mechanisms, they are typically presented as individual
end-to-end designs. To our knowledge, prior work does not provide a shared
abstraction for reasoning about which combinations have been explored and
which remain uncovered within a clearly stated policy space.

We find that the surveyed designs can be factored into two
independently specifiable dimensions: \emph{expert residency} and
\emph{MoE execution strategy}. Expert residency determines which experts occupy
GPU memory when a layer is reached and how the resident set is updated for
subsequent layers. The MoE execution strategy determines where the experts
selected by the current layer are computed and how computation, weight
transfers, and synchronization are coordinated. Combining one choice from each
dimension defines a concrete candidate within the declared space.

The two decisions are separable as policy choices because they govern
different points in the decode sequence. A layer begins expert execution with
its residency state already determined; prediction and cache updates launched
alongside that computation prepare later layers. Execution can therefore use
the current resident set without constraining how it was chosen. Thus $r$
residency choices and $e$ execution choices yield $re$ combinations from
$r+e$ reusable modules, although shared resource use still requires evaluating
each combination's timing.

\textbf{Scope of the Policy Space.}
The Policy Matrix is further restricted to policies designed to preserve model
output quality. Each candidate follows the model's original routing decisions,
executes every selected expert, and retains the original expert precision.
Methods that prune or dynamically skip experts
\cite{not-all-experts-equal}, substitute a router-selected expert with a cached
alternative \cite{smoe}, or execute cache-miss experts at a lower precision
\cite{moe-apex} alter either the set of executed experts or their numerical
representation. Such methods therefore fall outside the scope of the current
Policy Matrix.

\section{Design}
\label{sec:method}
\label{sec:policy-space}

Observation~I shows that the fastest offloading policy depends on both the
model and the platform. COPE therefore separates policy specification from
configuration-specific performance prediction
(Figure~\ref{fig:zcache-overview}). The Policy Matrix factors monolithic
offloading designs into two independently specifiable choices: \emph{expert
residency}, which determines which experts occupy GPU memory and how the cache
is updated for future layers; and the \emph{MoE execution strategy}, which
determines where and how the current layer's routed experts execute. The
\emph{Virtual Heterogeneous Machine} (VHM) then estimates the TPOT of the
resulting candidates without running each candidate end to end. It profiles
atomic operators and short combined-operator schedules on the target platform,
while an offline, timing-free routing trace records the model's routed and
predicted expert IDs. VHM replays each candidate's
cache states using the trace and composes the measured costs into
a whole-model TPOT estimate.

\definecolor{zoblue}{HTML}{8FAADC}
\definecolor{zoorange}{HTML}{F4B183}
\definecolor{zogreen}{HTML}{548235}
\definecolor{zolightgreen}{HTML}{C6E0B4}
\definecolor{zogold}{HTML}{FFD966}
\definecolor{zonavy}{HTML}{1F3864}
\definecolor{zoselected}{HTML}{7F7F7F}
\definecolor{zoprofile}{HTML}{4472C4}
\definecolor{zogray}{HTML}{A6A6A6}

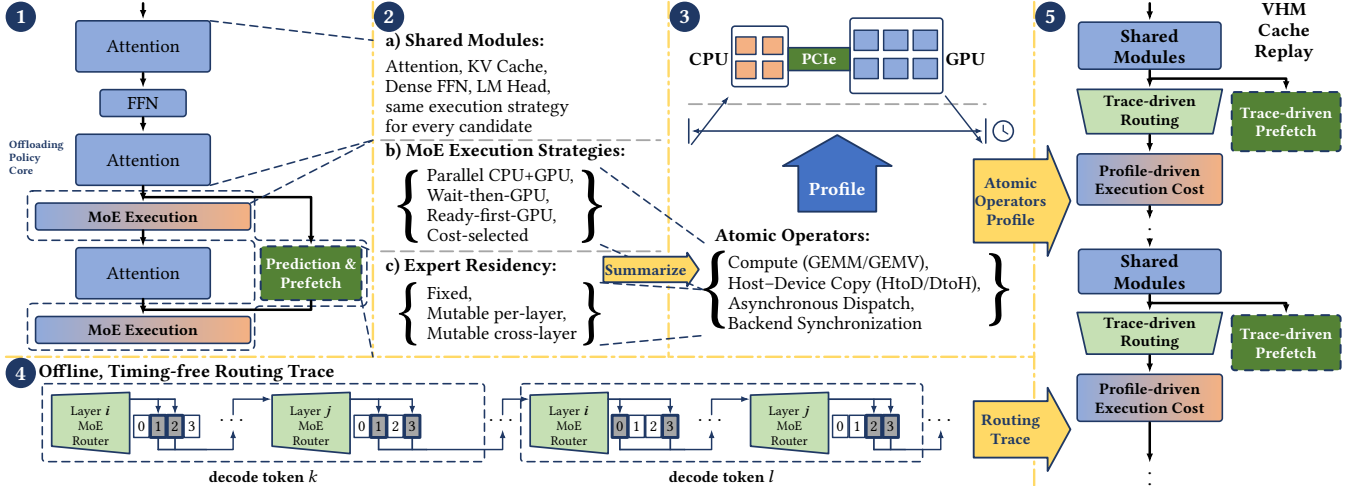
\begin{figure*}[t]
  \centering
  \setlength{\belowcaptionskip}{-5pt}
  \begin{tikzpicture}[
      x=\dimexpr\textwidth/100\relax,
      y=-\dimexpr\textwidth/100\relax,
      line cap=round,line join=round,
      zotext/.style={font=\scriptsize,inner sep=0pt,text=black},
      zosmall/.style={font=\tiny,inner sep=0pt,text=black},
      zobox/.style={draw=zonavy,line width=0.6pt,rounded corners=1.2pt},
      zoarr/.style={-{Latex[length=1.15mm,width=0.8mm]},draw=zonavy,line width=0.55pt},
      zoflow/.style={-{Latex[length=1.25mm,width=0.9mm]},draw=black,line width=0.95pt},
      zobranch/.style={draw=black,line width=0.95pt},
      zosoft/.style={draw=zogray,line width=0.7pt,dash pattern=on 5pt off 2.4pt},
      zoassoc/.style={draw=zonavy,line width=0.6pt,dash pattern=on 3pt off 2pt},
      zomodule/.style={draw=zogold,line width=0.9pt,
        dash pattern=on 6pt off 2pt on 1pt off 2pt},
    ]
    \useasboundingbox (0,0) rectangle (100,36.1);

    \draw[zomodule] (27.3,0) -- (27.3,26.4);
    \draw[zomodule] (49.3,0) -- (49.3,26.4);
    \draw[zomodule] (76.5,0) -- (76.5,36.1);
    \draw[zomodule] (0,26.4) -- (76.5,26.4);
    \draw[zosoft] (27.9,10.28) -- (48.7,10.28);
    \draw[zosoft] (27.9,18.55) -- (48.7,18.55);
    \draw[zosoft] (50.8,7.6) -- (72.9,7.6);

    \foreach \n/\x/\y in {1/1.1/1.1,2/28.5/1.1,3/50.5/1.1,4/1.1/27.6,5/77.7/1.1}{
      \node[draw=zonavy,fill=zonavy,text=white,circle,line width=0.6pt,
        minimum size=3.8mm,inner sep=0pt,font=\fontsize{8}{8}\selectfont\bfseries]
        at (\x,\y) {\n};
    }

    \draw[zoassoc,rounded corners=2pt]
      (1.6,14.05) -- (18.5,14.05) -- (18.5,17.73) -- (26.85,17.73)
      -- (26.85,22.63) -- (18.5,22.63) -- (18.5,25.90) -- (1.6,25.90)
      -- (1.6,22.43) -- (18.5,22.43) -- (18.5,17.73) -- (1.6,17.73) -- cycle;
    \node[anchor=north west,align=left,font=\fontsize{4.2}{4.8}\selectfont\bfseries,
      text=zonavy,inner sep=0pt] at (0.2,10.2) {Offloading\\Policy\\Core};

    \path[zobox,fill=zoblue] (5.2,1.40) rectangle (15.2,5.20);
    \node[zotext] at (10.2,3.30) {Attention};
    \path[zobox,fill=zoblue] (7.0,6.52) rectangle (13.4,8.52);
    \node[zotext] at (10.2,7.52) {FFN};
    \path[zobox,fill=zoblue] (5.2,9.84) rectangle (15.2,13.64);
    \node[zotext] at (10.2,11.74) {Attention};

    \path[shade,left color=zoblue,right color=zoorange,rounded corners=1.2pt]
      (2.2,14.96) rectangle (18.2,16.96);
    \draw[zobox,fill=none] (2.2,14.96) rectangle (18.2,16.96);
    \node[zosmall,font=\tiny\bfseries] at (10.2,15.96) {MoE Execution};

    \path[zobox,fill=zoblue] (5.2,18.28) rectangle (15.2,22.08);
    \node[zotext] at (10.2,20.18) {Attention};
    \path[zobox,fill=zogreen,dashed,line width=0.75pt]
      (18.95,18.18) rectangle (26.45,22.18);
    \node[align=center,font=\tiny\bfseries,text=white,inner sep=0pt]
      at (22.70,20.18) {Prediction \&\\Prefetch};

    \path[shade,left color=zoblue,right color=zoorange,rounded corners=1.2pt]
      (2.2,23.40) rectangle (18.2,25.40);
    \draw[zobox,fill=none] (2.2,23.40) rectangle (18.2,25.40);
    \node[zosmall,font=\tiny\bfseries] at (10.2,24.40) {MoE Execution};

    \draw[zoflow] (10.2,0.1) -- (10.2,1.3);
    \draw[zoflow] (10.2,5.3) -- (10.2,6.4);
    \draw[zoflow] (10.2,8.6) -- (10.2,9.7);
    \draw[zoflow] (10.2,13.7) -- (10.2,14.9);
    \draw[zoflow] (10.2,17.0) -- (10.2,18.2);
    \draw[zoflow] (10.2,22.1) -- (10.2,23.3);
    \draw[zoflow] (10.2,14.52) -- (22.70,14.52) -- (22.70,18.10);
    \draw[zobranch] (22.70,22.26) -- (22.70,22.90) -- (10.2,22.90);

    \draw[zoassoc] (15.2,1.40) -- (27.3,2.85);
    \draw[zoassoc] (15.2,13.64) -- (27.3,10.28);
    \draw[zoassoc] (18.2,14.96) -- (27.3,10.28);
    \draw[zoassoc] (26.45,18.18) -- (27.3,18.55);
    \draw[zoassoc] (26.45,22.18) -- (27.3,26.4);

    \node[anchor=west,font=\scriptsize\bfseries,inner sep=0pt]
      at (28.2,2.9) {a) Shared Modules:};
    \node[anchor=north west,align=left,font=\fontsize{7}{7.6}\selectfont,inner sep=0pt]
      at (28.2,4.2) {Attention, KV Cache,\\Dense FFN, LM Head,\\same execution strategy\\for every candidate};

    \node[anchor=west,font=\scriptsize\bfseries,inner sep=0pt]
      at (28.2,11.1) {b) MoE Execution Strategies:};
    \node[font=\fontsize{30}{30}\selectfont,inner sep=0pt] at (29.9,15.2) {\{};
    \node[anchor=north west,align=left,font=\fontsize{7}{7.6}\selectfont,inner sep=0pt]
      at (31.3,12.3) {Parallel CPU+GPU,\\Wait-then-GPU,\\Ready-first-GPU,\\Cost-selected};
    \node[font=\fontsize{30}{30}\selectfont,inner sep=0pt] at (43.6,15.2) {\}};

    \node[anchor=west,font=\scriptsize\bfseries,inner sep=0pt]
      at (28.2,19.7) {c) Expert Residency:};
    \node[font=\fontsize{26}{26}\selectfont,inner sep=0pt] at (29.9,23.2) {\{};
    \node[anchor=north west,align=left,font=\fontsize{7}{7.6}\selectfont,inner sep=0pt]
      at (31.3,21.1) {Fixed,\\Mutable per-layer,\\Mutable cross-layer};
    \node[font=\fontsize{26}{26}\selectfont,inner sep=0pt] at (43.6,23.2) {\}};

    \draw[zoassoc] (44.24,12.35) -- (52.05,18.00);
    \draw[zoassoc] (44.24,18.05) -- (51.85,21.40);
    \draw[zoassoc] (44.22,20.88) -- (51.85,21.40);
    \draw[zoassoc] (44.22,25.52) -- (52.05,24.80);
    \path[fill=zogold,draw=zonavy,line width=0.7pt]
      (44.45,18.95) -- (50.68,18.95) -- (50.68,18.65) -- (51.55,19.90)
      -- (50.68,21.15) -- (50.68,20.85) -- (44.45,20.85) -- cycle;
    \node[font=\tiny\bfseries,text=zonavy,inner sep=0pt]
      at (47.60,19.90) {Summarize};

    \node[anchor=west,font=\scriptsize\bfseries,inner sep=0pt] at (50.8,4.2) {CPU};
    \draw[zobox] (54.0,1.8) rectangle (58.2,6.3);
    \foreach \xx/\yy in {54.3/2.6,56.3/2.6,54.3/4.4,56.3/4.4}
      \path[draw=zonavy,fill=zoorange,line width=0.55pt]
        (\xx,\yy) rectangle ++(1.5,1.3);

    \path[draw=zonavy,fill=zogreen,line width=0.6pt]
      (58.2,3.4) rectangle (62.8,5.1);
    \node[font=\tiny\bfseries,text=white,inner sep=0pt] at (60.5,4.25) {PCIe};

    \draw[zobox] (62.8,1.2) rectangle (69.7,6.8);
    \foreach \xx in {63.1,65.3,67.5}
      \foreach \yy in {2.1,4.2}
        \path[draw=zonavy,fill=zoblue,line width=0.55pt]
          (\xx,\yy) rectangle ++(1.8,1.5);
    \node[anchor=east,font=\scriptsize\bfseries,inner sep=0pt] at (72.9,4.2) {GPU};

    \draw[zoarr] (51.0,10.6) -- (53.8,7.0);
    \draw[zoarr] (69.9,7.0) -- (72.7,10.6);
    \draw[draw=zonavy,line width=0.6pt] (50.8,8.7) -- (50.8,10.5);
    \draw[draw=zonavy,line width=0.6pt] (72.9,8.7) -- (72.9,10.5);
    \draw[{Latex[length=1.0mm]}-{Latex[length=1.0mm]},draw=zonavy,line width=0.6pt]
      (51.0,9.6) -- (72.7,9.6);
    \draw[draw=zonavy,line width=0.6pt,fill=white] (74.2,9.6) circle (0.75);
    \draw[draw=zonavy,line width=0.5pt] (74.2,9.6) -- ++(0,-0.43);
    \draw[draw=zonavy,line width=0.5pt] (74.2,9.6) -- ++(0.34,0.22);

    \path[fill=zoprofile,draw=zonavy,line width=0.7pt]
      (58.95,15.6) -- (64.75,15.6) -- (64.75,12.2) -- (65.9,12.2)
      -- (61.85,9.9) -- (57.8,12.2) -- (58.95,12.2) -- cycle;
    \node[font=\scriptsize\bfseries,text=white,inner sep=0pt]
      at (61.85,13.9) {Profile};

    \node[anchor=west,font=\scriptsize\bfseries,inner sep=0pt]
      at (52.7,17.5) {Atomic Operators:};
    \node[font=\fontsize{32}{32}\selectfont,inner sep=0pt] at (52.7,21.4) {\{};
    \node[anchor=north west,align=left,font=\fontsize{7}{7.6}\selectfont,inner sep=0pt]
      at (53.7,18.80) {Compute (GEMM/GEMV),\\Host--Device Copy (HtoD/DtoH),\\
        Asynchronous Dispatch,\\Backend Synchronization};
    \node[font=\fontsize{30}{30}\selectfont,inner sep=0pt] at (73.9,21.4) {\}};

    \path[fill=zogold,draw=zonavy,line width=0.7pt]
      (72.0,11.7) -- (77.5,11.7) -- (77.5,10.7) -- (79.2,14.8)
      -- (77.5,18.9) -- (77.5,17.9) -- (72.0,17.9) -- cycle;
    \node[align=center,font=\tiny\bfseries,text=zonavy,inner sep=0pt]
      at (74.75,14.8) {Atomic\\Operators\\Profile};

    \node[anchor=west,font=\scriptsize\bfseries,inner sep=0pt]
      at (2.4,27.55) {Offline, Timing-free Routing Trace};
    \draw[zoassoc,rounded corners=1.2pt] (2.7,28.4) rectangle (35.6,34.2);
    \draw[zoassoc,rounded corners=1.2pt] (38.3,28.4) rectangle (68.2,34.2);

    \foreach \x/\layer/\selA/\selB in {
        3.4/i/1/2,19.8/j/1/3,39.0/i/0/3,55.4/j/2/3}{
      \path[draw=zonavy,fill=zolightgreen,line width=0.65pt]
        (\x,29.25) -- (\x+5.8,29.05) -- (\x+5.8,33.60)
        -- (\x,33.05) -- cycle;
      \node[align=center,font=\fontsize{5}{5.5}\selectfont,inner sep=0pt]
        at (\x+2.90,31.52) {Layer $\layer$\\MoE\\Router};
      \foreach \eid in {0,1,2,3}{
        \pgfmathsetmacro{\tx}{\x+6.1+1.25*\eid}
        \ifnum\eid=\selA
          \path[draw=zonavy,line width=1.1pt,fill=zogray]
            (\tx,30.75) rectangle ++(1.15,1.65);
        \else\ifnum\eid=\selB
          \path[draw=zonavy,line width=1.1pt,fill=zogray]
            (\tx,30.75) rectangle ++(1.15,1.65);
        \else
          \path[draw=zonavy,line width=0.55pt,fill=white]
            (\tx,30.75) rectangle ++(1.15,1.65);
        \fi\fi
        \node[font=\tiny,inner sep=0pt] at (\tx+0.575,31.575) {\eid};
      }
      \pgfmathsetmacro{\sa}{\x+6.1+1.25*\selA+0.575}
      \pgfmathsetmacro{\sb}{\x+6.1+1.25*\selB+0.575}
      \pgfmathsetmacro{\smax}{max(\sa,\sb)}
      \pgfmathsetmacro{\smin}{min(\sa,\sb)}
      \draw[draw=zonavy,line width=0.65pt] (\x+5.8,29.55) -- (\smax,29.55);
      \draw[zoarr] (\sa,29.55) -- (\sa,30.70);
      \draw[zoarr] (\sb,29.55) -- (\sb,30.70);
      \draw[draw=zonavy,line width=0.65pt] (\sa,32.45) -- (\sa,33.30);
      \draw[draw=zonavy,line width=0.65pt] (\sb,32.45) -- (\sb,33.30);
      \draw[draw=zonavy,line width=0.65pt] (\smin,33.30) -- (\x+11.50,33.30);
    }

    \draw[zoarr] (14.90,33.30) -- (16.90,33.30) -- (16.90,31.80);
    \node[font=\scriptsize\bfseries,inner sep=0pt] at (16.90,31.25) {$\cdots$};
    \draw[zoarr] (16.90,30.70) -- (16.90,29.55) -- (19.75,29.55);
    \draw[zoarr] (31.30,33.30) -- (36.95,33.30) -- (36.95,31.80);

    \node[font=\scriptsize\bfseries,inner sep=0pt] at (36.95,31.25) {$\cdots$};
    \draw[zoarr] (36.95,30.70) -- (36.95,29.55) -- (38.95,29.55);

    \draw[zoarr] (50.50,33.30) -- (52.50,33.30) -- (52.50,31.80);
    \node[font=\scriptsize\bfseries,inner sep=0pt] at (52.50,31.25) {$\cdots$};
    \draw[zoarr] (52.50,30.70) -- (52.50,29.55) -- (55.35,29.55);
    \draw[zoarr] (66.90,33.30) -- (69.55,33.30) -- (69.55,31.80);
    \node[font=\scriptsize\bfseries,inner sep=0pt] at (69.55,31.25) {$\cdots$};

    \node[font=\tiny\bfseries,inner sep=0pt] at (19.15,35.25) {decode token $k$};
    \node[font=\tiny\bfseries,inner sep=0pt] at (53.25,35.25) {decode token $l$};

    \path[fill=zogold,draw=zonavy,line width=0.7pt]
      (72.0,29.7) -- (77.5,29.7) -- (77.5,28.7) -- (79.2,31.8)
      -- (77.5,34.9) -- (77.5,33.9) -- (72.0,33.9) -- cycle;
    \node[align=center,font=\tiny\bfseries,text=zonavy,inner sep=0pt]
      at (74.75,31.8) {Routing\\Trace};

    \node[align=center,font=\scriptsize\bfseries,inner sep=0pt]
      at (95.1,2.25) {VHM\\Cache\\Replay};

    \draw[zoflow] (85.1,0.1) -- (85.1,1.4);
    \path[zobox,fill=zoblue] (79.8,1.5) rectangle (90.4,5.0);
    \node[align=center,font=\scriptsize\bfseries,inner sep=0pt]
      at (85.1,3.25) {Shared\\Modules};
    \path[draw=zonavy,fill=zolightgreen,line width=0.65pt]
      (79.7,6.5) -- (90.5,6.5) -- (89.8,9.8) -- (80.4,9.8) -- cycle;
    \node[align=center,font=\tiny\bfseries,inner sep=0pt]
      at (85.1,8.15) {Trace-driven\\Routing};
    \path[shade,left color=zoblue,right color=zoorange,rounded corners=1.2pt]
      (79.7,11.3) rectangle (90.5,15.1);
    \draw[zobox,fill=none] (79.7,11.3) rectangle (90.5,15.1);
    \node[align=center,font=\tiny\bfseries,inner sep=0pt]
      at (85.1,13.2) {Profile-driven\\Execution Cost};

    \path[zobox,fill=zogreen,dashed,line width=0.75pt]
      (91.2,6.7) rectangle (99.3,11.1);
    \node[align=center,font=\tiny\bfseries,text=white,inner sep=0pt]
      at (95.25,8.9) {Trace-driven\\Prefetch};

    \draw[zoflow] (85.1,5.1) -- (85.1,6.4);
    \draw[zoflow] (85.1,9.9) -- (85.1,11.2);
    \draw[zoflow] (85.1,15.2) -- (85.1,16.2);
    \draw[zoflow] (85.1,5.7) -- (95.25,5.7) -- (95.25,6.6);
    \node[font=\scriptsize\bfseries,inner sep=0pt] at (85.1,17.1) {$\vdots$};

    \path[zobox,fill=zoblue] (79.8,18.4) rectangle (90.4,21.7);
    \node[align=center,font=\scriptsize\bfseries,inner sep=0pt]
      at (85.1,20.05) {Shared\\Modules};
    \path[draw=zonavy,fill=zolightgreen,line width=0.65pt]
      (79.7,23.1) -- (90.5,23.1) -- (89.8,26.2) -- (80.4,26.2) -- cycle;
    \node[align=center,font=\tiny\bfseries,inner sep=0pt]
      at (85.1,24.65) {Trace-driven\\Routing};
    \path[shade,left color=zoblue,right color=zoorange,rounded corners=1.2pt]
      (79.7,27.7) rectangle (90.5,31.2);
    \draw[zobox,fill=none] (79.7,27.7) rectangle (90.5,31.2);
    \node[align=center,font=\tiny\bfseries,inner sep=0pt]
      at (85.1,29.45) {Profile-driven\\Execution Cost};

    \path[zobox,fill=zogreen,dashed,line width=0.75pt]
      (91.2,23.3) rectangle (99.3,27.4);
    \node[align=center,font=\tiny\bfseries,text=white,inner sep=0pt]
      at (95.25,25.35) {Trace-driven\\Prefetch};

    \draw[zoflow] (85.1,17.8) -- (85.1,18.3);
    \draw[zoflow] (85.1,21.8) -- (85.1,23.0);
    \draw[zoflow] (85.1,26.3) -- (85.1,27.6);
    \draw[zoflow] (85.1,22.4) -- (95.25,22.4) -- (95.25,23.2);
    \draw[zoflow] (85.1,31.3) -- (85.1,34.3);
    \node[font=\scriptsize\bfseries,inner sep=0pt] at (85.1,35.25) {$\vdots$};

  \end{tikzpicture}%
  \caption{Overview of COPE.
  The routing trace contains routed and predicted expert IDs; the panel illustrates routed IDs across decode steps.
  The atomic-operator profile and routing trace are the only inputs to VHM cache replay.}
  \label{fig:zcache-overview}
\end{figure*}

\subsection{Modularizing Offloading Policies with the Policy Matrix}
\label{sec:policy-reduction}
\label{sec:policy-dimensions}

Existing MoE offloaders are typically presented as monolithic designs, making
the space of alternatives difficult to navigate. On a new model--hardware
configuration, finding the fastest design would require implementing and
testing each existing system. Even then, an unordered list of systems offers
little indication of which promising combinations have yet to be explored.

The Policy Matrix exposes two decisions within these designs: \emph{expert
residency} and \emph{MoE execution strategy}. Residency determines which
experts occupy GPU memory when a layer is reached, and whether to update that
cache in anticipation of later layers. Execution determines where the experts
selected by the current layer's router run, and how their computation and
weight movement are coordinated.

\definecolor{hotblue}{RGB}{111,147,201}
\definecolor{coldorange}{RGB}{249,214,173}
\definecolor{copygreen}{RGB}{84,130,53}
\definecolor{schedyellow}{RGB}{255,217,102}
\definecolor{htodgreen}{RGB}{75,115,48}

\begin{figure*}[t]
  \centering
  \setlength{\belowcaptionskip}{-5pt}
  \begin{tikzpicture}[
      x=\dimexpr\textwidth/18\relax,
      y=-\dimexpr\textwidth/18\relax,
      line cap=round,line join=round,
      pmgrid/.style={draw=black,line width=0.6pt},
      pmarr/.style={-{Latex[length=1.15mm,width=0.8mm]},draw=black,line width=0.55pt},
      pmcopy/.style={-{Latex[length=1.1mm,width=0.8mm]},draw=copygreen,line width=0.6pt},
      pmhot/.style={draw=black,line width=0.65pt,fill=hotblue},
      pmcold/.style={draw=black,dashed,line width=0.5pt,fill=coldorange},
      pmstage/.style={draw=black,line width=0.55pt,fill=htodgreen},
      pmmerge/.style={draw=black,circle,line width=0.55pt,fill=white,
        minimum size=0.27cm,inner sep=0pt,font=\scriptsize},
      pmtxt/.style={font=\scriptsize,inner sep=0pt},
    ]

    \node[font=\Large,inner sep=0pt] at (2.45,0.37) {\{};
    \node[anchor=west,pmtxt] (policyset) at (2.68,0.37)
      {Offloading Policy 1,\quad Offloading Policy 2,\quad Offloading Policy 3,\quad ...};
    \node[font=\Large,inner sep=0pt] at ([xshift=7.2pt]policyset.east) {\}};
    \fill[black!65] (5.32,0.61) -- (5.80,0.61) -- (5.80,0.73)
      -- (6.14,0.73) -- (5.56,0.86) -- (4.98,0.73) -- (5.32,0.73) -- cycle;
    \node[anchor=north west,font=\footnotesize\bfseries,inner sep=0pt]
      at (11.24,0.08) {Policy Set (Naive, Unstructured)};
    \fill[black!65] (11.90,0.365) -- (12.30,0.365) -- (12.30,0.445)
      -- (12.60,0.445) -- (12.10,0.535) -- (11.60,0.445)
      -- (11.90,0.445) -- cycle;
    \node[anchor=south west,font=\footnotesize\bfseries,inner sep=0pt]
      at (11.24,0.84) {Policy Matrix (Composable, Structured)};

    \draw[pmgrid] (0,0.88) rectangle (17.8,5.12);
    \foreach \xx in {3.60,7.15,10.70,14.25}
      \draw[pmgrid] (\xx,0.88) -- (\xx,5.12);
    \foreach \yy in {2.28,3.25,4.13}
      \draw[pmgrid] (0,\yy) -- (17.8,\yy);
    \draw[pmgrid] (0,0.88) -- (3.60,2.28);

    \node[anchor=north east,align=right,font=\scriptsize\bfseries,inner sep=0pt]
      at (3.40,1.04) {Execution\\Strategy};
    \node[anchor=south west,font=\scriptsize\bfseries,inner sep=0pt]
      at (0.18,2.13) {Residency};

    \node[pmtxt] at (5.375,1.08) {Parallel CPU+GPU};
    \node[pmtxt] at (8.925,1.08) {Wait-then-GPU};
    \node[pmtxt] at (12.475,1.08) {Ready-first GPU};
    \node[pmtxt] at (16.025,1.08) {Cost-selected};

    \begin{scope}[shift={(3.84,1.42)}]
      \draw[pmarr] (0.02,0.40) -- (0.38,0.40) -- (0.38,0.17) -- (0.56,0.17);
      \draw[pmarr] (0.38,0.40) -- (0.38,0.66) -- (0.56,0.66);
      \path[pmhot] (0.56,0.01) rectangle (1.31,0.33);
      \node[pmtxt] at (0.935,0.17) {GPU};
      \path[pmcold] (0.56,0.50) rectangle (1.72,0.82);
      \node[pmtxt] at (1.14,0.66) {CPU};
      \node[pmmerge] (m) at (2.55,0.40) {$+$};
      \draw[pmarr] (1.31,0.17) -- (2.25,0.17) -- (m.north);
      \draw[pmarr] (1.72,0.66) -- (2.25,0.66) -- (m.south);
      \draw[pmarr] (m.east) -- (2.97,0.40);
    \end{scope}

    \begin{scope}[shift={(7.37,1.42)}]
      \draw[pmarr] (0.02,0.34) -- (0.40,0.34) -- (0.40,0.18) -- (0.52,0.18);
      \path[pmhot] (0.52,0.02) rectangle (1.19,0.34);
      \node[pmtxt] at (0.855,0.18) {GPU};
      \node[pmmerge] (m) at (1.50,0.18) {$+$};
      \path[pmstage] (1.25,0.53) rectangle (2.18,0.81);
      \node[pmtxt,text=white] at (1.715,0.67) {HtoD};
      \path[pmhot] (2.34,0.02) rectangle (3.01,0.34);
      \node[pmtxt] at (2.675,0.18) {GPU};
      \draw[pmarr] (1.19,0.18) -- (m.west);
      \draw[pmarr] (m.east) -- (2.34,0.18);
      \draw[pmarr] (1.19,0.31) -- (1.25,0.67);
      \draw[pmarr] (2.18,0.67) -- (2.34,0.31);
    \end{scope}

    \begin{scope}[shift={(10.92,1.42)}]
      \draw[pmarr] (0.02,0.40) -- (0.40,0.40) -- (0.40,0.18) -- (0.52,0.18);
      \draw[pmarr] (0.40,0.40) -- (0.40,0.67) -- (0.52,0.67);
      \path[pmhot] (0.52,0.02) rectangle (1.19,0.34);
      \node[pmtxt] at (0.855,0.18) {GPU};
      \node[pmmerge] (m) at (1.50,0.18) {$+$};
      \path[pmstage] (0.52,0.53) rectangle (2.18,0.81);
      \node[pmtxt,text=white] at (1.35,0.67) {HtoD};
      \path[pmhot] (2.34,0.02) rectangle (3.01,0.34);
      \node[pmtxt] at (2.675,0.18) {GPU};
      \draw[pmarr] (1.19,0.18) -- (m.west);
      \draw[pmarr] (m.east) -- (2.34,0.18);
      \draw[pmarr] (2.18,0.67) -- (2.34,0.31);
    \end{scope}

    \begin{scope}[shift={(14.46,1.37)}]
      \node[diamond,draw=black,fill=schedyellow,line width=0.55pt,
        minimum width=0.43cm,minimum height=0.43cm,inner sep=0pt] (d)
        at (0.27,0.41) {};
      \draw[pmarr] (0.00,0.41) -- (d.west);
      \draw[pmarr] (d.east) -- (0.64,0.41) -- (0.64,0.10) -- (0.78,0.10);
      \draw[pmarr] (0.64,0.41) -- (0.78,0.41);
      \draw[pmarr] (0.64,0.41) -- (0.64,0.76) -- (0.78,0.76);
      \path[pmhot] (0.78,-0.06) rectangle (1.37,0.26);
      \node[pmtxt] at (1.075,0.10) {GPU};
      \node[pmmerge] (m1) at (1.64,0.10) {$+$};
      \path[pmstage] (0.78,0.335) rectangle (2.03,0.565);
      \node[pmtxt,text=white] at (1.405,0.45) {HtoD};
      \path[pmcold] (0.78,0.64) rectangle (2.40,0.88);
      \node[pmtxt] at (1.59,0.76) {CPU};
      \path[pmhot] (2.18,-0.06) rectangle (2.76,0.26);
      \node[pmtxt] at (2.47,0.10) {GPU};
      \node[pmmerge] (m2) at (3.02,0.41) {$+$};
      \draw[pmarr] (1.37,0.10) -- (m1.west);
      \draw[pmarr] (m1.east) -- (2.18,0.10);
      \draw[pmarr] (2.03,0.45) -- (2.18,0.20);
      \draw[pmarr] (2.76,0.10) -- (2.88,0.10) -- (m2.north);
      \draw[pmarr] (2.40,0.76) -- (2.88,0.76) -- (m2.south);
      \draw[pmarr] (m2.east) -- (3.26,0.41);
    \end{scope}

    \node[anchor=west,font=\scriptsize\bfseries,inner sep=0pt]
      at (0.17,2.48) {Fixed};
    \node[anchor=north west,align=left,font=\scriptsize\itshape,
      text=black!70,inner sep=0pt]
      at (0.17,2.70) {fill by index or \\popularity-order};
    \begin{scope}[shift={(2.15,2.46)},xscale=0.74]
      \foreach \yy in {0,0.36} {
        \path[pmhot] (0,\yy) rectangle (0.52,\yy+0.16);
        \path[pmcold] (0.52,\yy) rectangle (1.42,\yy+0.16);
        \path[pattern=north east lines,pattern color=black] (0.18,\yy) rectangle (0.31,\yy+0.16);
        \path[pattern=north east lines,pattern color=black] (1.03,\yy) rectangle (1.16,\yy+0.16);
      }
      \draw[pmcopy] (1.09,0.16) .. controls (1.02,0.44) and (0.54,0.64) .. (0.25,0.52);
      \draw[pmcopy] (1.09,0.52) .. controls (1.02,0.24) and (0.54,0.04) .. (0.25,0.16);
      \draw[red!80!black,line width=1pt] (0.63,0.25) -- (0.83,0.48);
      \draw[red!80!black,line width=1pt] (0.83,0.25) -- (0.63,0.48);
      \node[pmtxt,anchor=west] at (1.47,0.08) {$L_j$};
      \node[pmtxt,anchor=west] at (1.47,0.44) {$L_i$};
    \end{scope}

    \node[anchor=north west,align=left,font=\scriptsize\bfseries,inner sep=0pt]
      at (0.17,3.43) {Mutable\\per-layer};
    \begin{scope}[shift={(2.15,3.38)},xscale=0.74]
      \foreach \yy in {0,0.36} {
        \path[pmhot] (0,\yy) rectangle (0.52,\yy+0.16);
        \path[pmcold] (0.52,\yy) rectangle (1.42,\yy+0.16);
        \path[pattern=north east lines,pattern color=black] (0.18,\yy) rectangle (0.31,\yy+0.16);
        \path[pattern=north east lines,pattern color=black] (1.03,\yy) rectangle (1.16,\yy+0.16);
      }
      \draw[pmcopy] (1.09,0.16) .. controls (1.04,0.32) and (0.51,0.32) .. (0.25,0.16);
      \draw[pmcopy] (1.09,0.52) .. controls (1.04,0.71) and (0.51,0.71) .. (0.25,0.52);
      \node[pmtxt,anchor=west] at (1.47,0.08) {$L_j$};
      \node[pmtxt,anchor=west] at (1.47,0.44) {$L_i$};
    \end{scope}

    \node[anchor=north west,align=left,font=\scriptsize\bfseries,inner sep=0pt]
      at (0.17,4.37) {Mutable\\cross-layer};
    \begin{scope}[shift={(2.15,4.43)},xscale=0.74]
      \foreach \yy in {0,0.36} {
        \path[pmhot] (0,\yy) rectangle (0.52,\yy+0.16);
        \path[pmcold] (0.52,\yy) rectangle (1.42,\yy+0.16);
        \path[pattern=north east lines,pattern color=black] (0.18,\yy) rectangle (0.31,\yy+0.16);
        \path[pattern=north east lines,pattern color=black] (1.03,\yy) rectangle (1.16,\yy+0.16);
      }
      \draw[pmcopy] (1.09,0.16) .. controls (1.02,0.44) and (0.54,0.64) .. (0.25,0.52);
      \draw[pmcopy] (1.09,0.52) .. controls (1.02,0.24) and (0.54,0.04) .. (0.25,0.16);
      \node[pmtxt,anchor=west] at (1.47,0.08) {$L_j$};
      \node[pmtxt,anchor=west] at (1.47,0.44) {$L_i$};
    \end{scope}

    \node[anchor=north west,align=left,font=\scriptsize,inner sep=0pt]
      at (3.77,2.47) {KTransformers~\cite{ktransformers}\\llama.cpp$^{a}$~\cite{llama-cpp}};
    \node[anchor=north west,align=left,font=\scriptsize,inner sep=0pt]
      at (7.31,2.62) {MoE-Lightning (batch-$n$)~\cite{moe-lightning}};
    \node[anchor=north west,align=left,font=\scriptsize,inner sep=0pt]
      at (10.86,2.62) {Discovered by COPE};
    \node[anchor=north west,align=left,font=\scriptsize,inner sep=0pt]
      at (14.41,2.62) {Fiddler~\cite{fiddler}};

    \node[anchor=north west,align=left,font=\scriptsize,inner sep=0pt]
      at (3.77,3.57) {Discovered by COPE};
    \node[anchor=north west,align=left,font=\scriptsize,inner sep=0pt]
      at (7.31,3.42) {Mixtral-Offloading~\cite{mixtral-offloading}\\DuoServe-MoE~\cite{duoserve-moe}};
    \node[anchor=north west,align=left,font=\scriptsize,inner sep=0pt]
      at (10.86,3.57) {ProMoE~\cite{promoe}};
    \node[anchor=north west,align=left,font=\scriptsize,inner sep=0pt]
      at (14.41,3.42) {HybriMoE~\cite{hybrimoe}\\DALI~\cite{dali}};

    \foreach \xx in {3.77,10.86}
      \node[anchor=north west,align=left,font=\scriptsize,inner sep=0pt]
        at (\xx,4.52) {Discovered by COPE};
    \node[anchor=north west,align=left,font=\scriptsize,inner sep=0pt]
      at (7.31,4.52) {MoE-Infinity$^{b}$~\cite{moe-infinity}};
    \node[anchor=north west,align=left,font=\scriptsize,inner sep=0pt]
      at (14.41,4.52) {FreeToken~\cite{freetoken}};
  \end{tikzpicture}%
  \vspace{4pt}

  \parbox{\textwidth}{\footnotesize\raggedright
    $^{a}$ MoE-only layer partition: $R_l=\varnothing$ or $E_l$; one expert branch per layer, separately evaluated.\\
    $^{b}$ Paper-level demand loading; resident-hit launch order is not established for the full wait-then-GPU execution sequence.\\
    \textit{Outside this matrix:} GPU remote reads (UVA)~\cite{vllm-uva};
    module prefetch~\cite{vllm-prefetch,sglang-offloaderv2}; approximate routing.}
  \caption{Policy Matrix: three expert-residency families crossed with four
  canonical MoE execution schedules. Under Fixed residency, the initial set may
  follow expert-ID or offline popularity order~\cite{fiddler}. Prior systems'
  workload and residency granularity may differ from COPE's batch-1, per-expert implementations.}

  \label{fig:policy-space}
\end{figure*}

\begin{figure*}[htp]
  \centering
  \includegraphics[width=\textwidth]{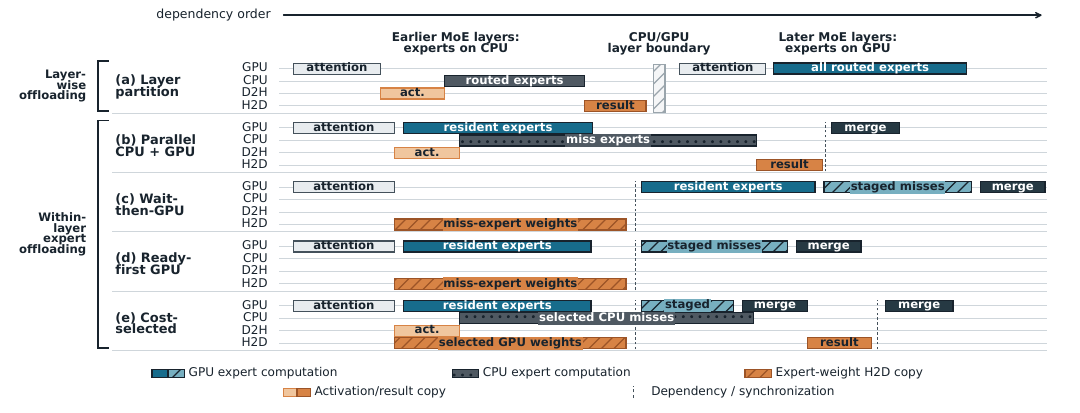}
  \caption{The layer-partition baseline and four cache execution schedules. Parallel sends misses to
  CPU; wait-then-GPU stages every miss; ready-first overlaps resident GPU work
  with staging; cost-selected assigns miss groups to CPU or staged-GPU
  execution. Prefetch is omitted.}

  \label{fig:execution-timelines}
\end{figure*}

Within the Policy Matrix, four execution strategies and three residency
families cover the mechanisms in our scope (Figure~\ref{fig:policy-space}).
Resident experts run on the GPU; a nonresident expert either runs on the CPU
and returns its result, or moves its complete weights to the GPU before
execution. \emph{Parallel} sends all misses to the CPU.
\emph{Wait-then-GPU} stages misses before GPU computation; \emph{ready-first}
overlaps resident computation with staging.
\emph{Cost-selected} uses estimated costs to divide misses between CPU and
staged GPU execution. On the residency axis, \emph{Fixed} retains its
initially loaded experts; \emph{Mutable per-layer} permits replacement within
each layer's budget; and \emph{Mutable cross-layer} permits replacement across
layers.
A Fixed cache may be filled in expert-ID order or in popularity order, ranked
by historical routing frequency~\cite{fiddler}; neither choice changes its
no-replacement rule during decode.
Figure~\ref{fig:execution-timelines} contrasts the execution orders: equal
cache-hit and miss counts need not imply equal TPOT when computation and
transfers overlap differently.

The resulting policy space is the product of residency and execution choices:
  $\mathcal P_{\mathrm{offload}}
  =\mathcal P_{\mathrm{residency}}\times\mathcal P_{\mathrm{execution}}$.
Mapping prior work onto this product reveals combinations not represented by
the surveyed systems. COPE instantiates the Fixed and
Mutable per-layer rows, yielding eight cache policies; the cross-layer row
remains a classified design family. Both instantiated cache families begin
with a popularity-ordered fill of each layer's expert slots. We evaluate
whole-layer partition separately as a ninth candidate. Because these
combinations share expert
computation and weight movement, VHM can profile their common operations and
compose policy estimates before each candidate is integrated as a production
offloader.

\label{sec:capability-contract}
\label{sec:policy-scope}

The matrix describes policies that preserve the model's computation during
batch-1 decode, even when only part of its expert set fits in VRAM. Each
candidate follows the checkpoint's router, executes every selected expert,
and retains its stored precision. Methods that change routing, expert
representation, or the set of executed experts fall outside this space
\cite{pregated-moe,moe-apex,qmoe,smoe}. COPE selects the supported candidate
with the lowest predicted TPOT, using a fixed order only to resolve an exact tie.

\subsection{Predicting Offloading Performance with the Virtual Heterogeneous Machine}

The Policy Matrix defines the candidates; VHM estimates their TPOT to select
the predicted winner for each model--hardware configuration. The two axes
organize the search, but building and testing every choice would still be
costly.
VHM instead works with reusable operations beneath those choices. Their
measured costs alone are not enough, however: a policy's latency also depends
on which experts the model routes to, which of them hit in the cache, and
which prefetches update later layers.

VHM reduces profiling to atomic operators shared across policy choices and
short combined-operator schedules that preserve their interactions. The
operators capture computation, data movement, and synchronization on the
target platform. Separating component costs from execution modeling also
underlies inference simulators such as Vidur, LLMServingSim~2.0, and
Frontier~\cite{vidur,llmservingsim2,frontier}, as well as training simulators
such as SimAI and ASTRA-sim~2.0~\cite{simai,astrasim2}. VHM applies this
approach to expert offloading, where cache residency and CPU/GPU execution
determine which operations run and how they interact. Our analytic baseline
adapts the Hierarchical Roofline Model (HRM) from
MoE-Lightning~\cite{moe-lightning}; direct measurements capture overheads
its estimates omit (Figure~\ref{fig:operator-accuracy}).
The combined-operator schedules then capture what isolated measurements cannot:
whether operations overlap, serialize, or contend when a policy executes.
This gives VHM policy-relevant costs without timing a production offloader.

VHM then uses an offline routing trace to drive a replay of each layer's expert cache,
turning those measured costs into a whole-model TPOT estimate. For each
layer, VHM compares the routed experts with the current cache to identify
hits and misses, then looks up the corresponding execution cost. For mutable
residency, traced predictions and the next layer's cache state likewise
determine whether a prefetch is needed and its associated cost. VHM combines
these costs with the work shared by all policies and aggregates them across
layers and traced tokens. Once the target's profile and the trace are given,
this composition needs no further hardware parameters; a new configuration
changes the measurements, not the replay.

\subsubsection{Abstracting Hardware with Atomic-Operator Profiles}
\label{sec:tpot-model}
\label{sec:calibration-interface}

VHM's first phase turns the Policy Matrix's structured choices into a
reusable profile of the target model and hardware. Timing every policy
end to end would bring back the cost of building every candidate. Instead,
VHM decomposes residency and execution into atomic operators shared across
choices (Figure~\ref{fig:zcache-overview}, parts 2--3). Unlike broader serving
and training simulators~\cite{vidur,llmservingsim2,frontier,simai,astrasim2},
VHM selects profiling units around the expert-offloading decision. CPU
execution of a cache miss, for instance, serves both Parallel and
cost-selected; staged GPU execution and expert-weight transfer serve
wait-then-GPU, ready-first, and
cost-selected. The profile therefore measures common ingredients rather than
each complete offloader.

\begin{figure}[t]
  \centering
  \includegraphics[width=\columnwidth]{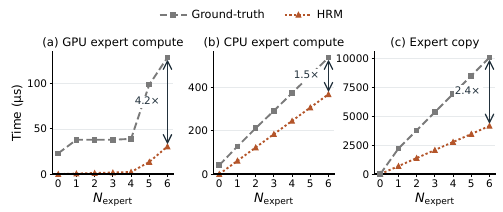}
  \caption{Measured atomic-operator latency and HRM estimates for
  DeepSeek-V2-Lite on RTX~4090 with 32 CPU threads.}
  \label{fig:operator-accuracy}
\end{figure}

VHM measures atomic-operator costs on the target instead of inferring them
solely from idealized compute and bandwidth limits. Our analytic baseline
adapts MoE-Lightning's Hierarchical Roofline Model (HRM)~\cite{moe-lightning}
to these operators. Figure~\ref{fig:operator-accuracy} shows why direct
measurement matters: CPU and GPU expert profiles incur nonzero costs even
with no routed experts, where the analytic estimate is zero. Expert staging
also includes preparation and readiness beyond weight transfer; at six
experts, its measured cost is $2.4\times$ the analytic estimate. Such probes
capture costs that a roofline bound omits while remaining reusable across
policies. Yet an accurate isolated cost is not necessarily an accurate
combined cost.
\label{sec:synthetic-probes}

To further capture interactions that isolated measurements miss, VHM also profiles
short combined-operator schedules. In ready-first, resident GPU work overlaps
expert staging; Parallel and cost-selected may overlap CPU and GPU expert
work. Wait-then-GPU instead holds GPU computation until staging finishes.
These dependencies, as well as contention and dispatch, can make the observed
duration differ from a sum or maximum of isolated costs. VHM measures the
relevant operations together, under their intended ordering and concurrency.
The resulting schedule costs, rather than an arithmetic assembly of isolated
timings, inform the subsequent TPOT estimate.

VHM profiles policy-invariant decode work once and reuses it for every
candidate. Attention, KV-cache work, dense FFNs, and output projection follow
the same execution path regardless of residency or MoE execution strategy.
Their shared profile varies with the target and context length, but not with
the offloading choice. Together with execution and prefetch schedule costs,
it provides the hardware-dependent inputs to trace-driven TPOT composition.

\subsubsection{Composing TPOT from Atomic Operator Profiles and Offline Traces}
\label{sec:policy-replay}

\begin{algorithm}[t]
\caption{VHM replay: predict one policy's TPOT from the profile and an
  offline routing trace.}
\label{alg:replay}
\begin{algorithmic}[1]
\Require \begin{tabular}[t]{@{}l@{}}
  trace: routed IDs $\mathrm{Trace}[t,l]$\\
  predicted next-layer IDs $\mathrm{Pred}[t,l]$\\
  profile: $\mathrm{Shared}$, $\mathrm{MoE}[e,h]$, $\mathrm{Prefetch}[r,q]$\\
  policy $p=(r,e)$: residency $r$, execution $e$
\end{tabular}
\Ensure predicted TPOT of $p$
\State $MC_l \gets$ cached experts of layer $l$
\State $\mathrm{acc} \gets 0$
\For{$t = 1$ to $T$}
  \State $\mathrm{acc} \gets \mathrm{acc} + \mathrm{Shared}(n)$ \Comment{shared work}
  \For{$l = 1$ to $L$}
    \State $h_l \gets |\mathrm{Trace}[t,l] \cap MC_l|$ \Comment{cache hits}
    \State $\mathrm{acc} \gets \mathrm{acc} + \mathrm{MoE}[e,h_l]$ \Comment{expert execution}
    \If{$r = \mathrm{Mutable\ per\text{-}layer}$ \textbf{and} $l < L$}
      \State $Q \gets \mathrm{Select}_r(\mathrm{Pred}[t,l] \setminus MC_{l+1})$
        \Comment{prefetch}
      \State $MC_{l+1} \gets \mathrm{Update}(MC_{l+1}, Q)$
      \State $\mathrm{acc} \gets \mathrm{acc} + \mathrm{Prefetch}[r,|Q|]$
    \EndIf
  \EndFor
\EndFor
\State \Return $\mathrm{acc}/T$ \Comment{TPOT}
\end{algorithmic}
\end{algorithm}

The offline routing trace contains both routed IDs (the experts actually routed to) and
predicted IDs (the experts predicted for the next layer), allowing VHM to estimate
cache-dependent TPOT. Routed IDs reveal which experts hit the current
layer's cache and thus which profiled execution cost applies. Predicted IDs
identify possible additions to the next layer's cache and the associated
prefetch cost; they need not match the experts later routed to that layer.
Both kinds of IDs are collected without timing a production offloader. Once
this trace and a target-measured profile are supplied, replay requires no
further hardware parameters.

VHM replays each candidate as a sequence of per-layer cache states
(Algorithm~\ref{alg:replay}). The cache policies start with each layer's
top-$K$ experts in popularity order; layer partition instead places every
expert in its later GPU-resident layers.
For each token and layer, it counts routed cache hits and retrieves the
execution cost for the candidate's strategy. Mutable residency also uses
the traced next-layer predictions to select at most one novel expert for
prefetch, updates that layer's cache, and accounts for prefetch cost.
\label{sec:nested-selection}
The compact cost lookups in the pseudocode stand for measured schedules,
including any CPU/GPU assignment within cost-selected; they do not add
isolated operator timings. Replay assumes a selected prefetch completes
before that layer is used, while prediction errors remain visible when
routed IDs differ from predicted IDs. Summing execution and prefetch costs
with shared decode work across tokens and layers gives the predicted TPOT.

\label{sec:calibration-validity}

A profile can be reused only while its target operating point remains valid;
trace changes and platform changes have different remedies
(Table~\ref{tab:validity}).

\begin{table}[t]
  \caption{Action required after an input changes.}
  \label{tab:validity}
  \centering
  \small
  \setlength{\tabcolsep}{3pt}
  \begin{tabular}{@{}lp{0.62\columnwidth}@{}}
    \toprule
    Action & Change \\
    \midrule
    Replay & Another candidate or a workload with an existing valid trace \\
    Retrace & Changed workload or predictor without a valid trace \\
    Recalibrate & Changed model or target operating point, including cache
      budget and compute/transfer resources \\
    Unsupported & Missing capabilities required by the candidate \\
    \bottomrule
  \end{tabular}
\end{table}

\section{Implementation}
\label{sec:validation-implementation}
For external validation, we implemented nine Policy Matrix candidates in
llama.cpp and evaluated their end-to-end paths on
RTX~4090, RTX~5090, and A100 platforms. This validation target uses the
ggml/CUDA backend~\cite{llama-cpp,ggml}. Separately, we implemented VHM's target
profiler, routing-trace collector, and profile-driven replay engine, encoding
the same nine candidates as execution schedules and cache-state transitions. We
evaluate VHM's prediction accuracy by comparing its candidate TPOT estimates
with end-to-end measurements from the llama.cpp paths.

The nine llama.cpp production paths serve only as validation and do not
contribute to VHM construction. In both replay and production, \emph{fixed}
and \emph{mutable residency} use the same per-layer popularity order and cache
capacity, and therefore start with identical resident experts. \emph{Fixed
residency} preserves this set, whereas \emph{mutable residency} updates it
through prediction and prefetch. The four execution strategies differ in their
handling of cache misses. \emph{Parallel} executes hits on the GPU and all
misses on the CPU. \emph{Wait-then-GPU} transfers all misses before GPU
execution, while \emph{ready-first} overlaps these transfers with resident GPU
work. \emph{Cost-selected} partitions misses between the CPU and GPU,
overlapping CPU execution, resident GPU work, and transfers before executing
the transferred experts. The eight cache candidates share a bounded pinned
buffer sized to concurrent routed work. \emph{Layer partition} instead
executes each complete MoE block on either the CPU or the GPU.

\subsection{Expert Routing Trace Collection}
\label{sec:trace-collection}

We collect expert-routing traces by replaying a set of representative
inputs through each MoE model. During policy-neutral decoding, a hook records
the experts selected by each layer's router. The trained next-layer expert
predictor also reads that layer's input hidden state and produces IDs for the
following layer; the same hook records these predictions. We pair the routed
and predicted IDs for each token and layer to form the routing trace used by
VHM. Routed IDs determine cache hits during replay, while predicted IDs drive
the possible updates to the next layer's cache.

\section{Evaluation}
\label{sec:evaluation}

We first test whether offloading winners change with the model and resources,
then assess winner prediction on A100 and two physical RTX devices. We report
absolute TPOT error, rank correlation, and top-1 selection separately. Top-1
means that predicted and measured minimum TPOT name the same candidate;
complexity and prediction intervals do not enter it. We also assess prediction
and calibration cost.

\subsection{Methodology}
\label{sec:evaluation-methodology}

\textbf{Platforms.}
We evaluate COPE on three GPU platforms: an A100-SXM4-80GB, an
RTX~4090, and an RTX~5090.
The A100-SXM4-80GB serves as the platform for our controlled resource
study. On the same machine, we vary the number of CPU workers, the
verified CUDA-MPS compute share~\cite{cuda-mps}, and the total-VRAM cap. These settings
provide sensitivity controls rather than emulating other GPUs.
The 24-GiB RTX~4090 is connected to two Xeon Gold 6430 CPUs over a
Gen4-x16-capable link. We evaluate one full-compute-share operating point
using the largest common feasible expert cache for each model, under an
admitted host-memory envelope of 128~GiB. The RTX~5090 platform provides
32~GiB of VRAM and 62.2~GiB of physical host memory. We evaluate 16-,
12-, and 8-GiB VRAM reservations, which leave capacity for another
application while retaining the full compute share. These reservations
do not emulate contention for SMs, host-memory bandwidth, or PCIe
bandwidth. We freeze CPU placement, memory limits, and the negotiated
interconnect for each physical-device cell.

\textbf{Models.}
We evaluate four MoE checkpoints: Qwen3.5-35B-A3B
\cite{qwen35,qwen35-35b-a3b-card}, Qwen3-30B-A3B
\cite{qwen3,qwen3-30b-a3b-card}, DeepSeek-Coder-V2-Lite-Instruct
\cite{deepseek-coder-v2,deepseek-coder-v2-lite-card}, and DeepSeek-V2-Lite-Chat
\cite{deepseek-v2,deepseek-v2-lite-chat-card}. The A100 experiments use Coder and the two
Qwen models, while the RTX~4090 experiments include all four models. The RTX~5090
experiments use Coder, Chat, and Qwen3 Q8\_0; Qwen-3.5 F16 is excluded because
its combined model and runtime footprint exceeds the available host-memory
budget.

\textbf{Baseline.}
We compare COPE with two estimators that replay identical routing traces and
cache states but estimate per-layer latency differently. Our \emph{atomic-operator
simulation}, inspired by profile-based simulators such as LLMServingSim~2.0~\cite{llmservingsim2},
composes independently measured operator costs using $\max$ for nominal overlap
and addition for serial execution. The \emph{Hierarchical
Roofline Model} (HRM) from MoE-Lightning~\cite{moe-lightning} instead derives
latency bounds from peak compute, memory, and interconnect throughput.
\emph{COPE} uses its Virtual Heterogeneous Machine to profile both atomic
operators and combined-operator schedules. End-to-end measurements from nine
llama.cpp implementations provide ground truth and are excluded from all
estimators.

\textbf{Workloads.}
We synthesize long-form prompts spanning dialogue, mathematical reasoning,
and software-engineering tasks, drawing on the domains of
ShareGPT~\cite{sharegpt}, MATH~\cite{math}, and
SWE-bench~\cite{swe-bench}. The evaluation prompts are disjoint from
predictor training. We measure each policy--prompt pair in five fresh
processes to reduce run-to-run variation in ground-truth TPOT.

\textbf{Predictor training.}
We collect model-generated traces on GSM8K~\cite{gsm8k},
Magicoder-Evol-Instruct-110K~\cite{magicoder-evol-instruct}, and UltraChat
200k~\cite{ultrachat-200k}. For each MoE layer, we record its input hidden
state and routed expert IDs. An MLP trained with PyTorch~\cite{pytorch} uses one layer's hidden state to
predict the next layer's routed IDs; training and validation are split
by prompt. We also derive each layer's cache popularity order from the
training datasets, without using evaluation prompts.

\subsection{Optimal Policy Diversity on A100}
\label{sec:policy-variation-evaluation}

\begin{figure}[t]
  \centering
  \includegraphics[width=\columnwidth]{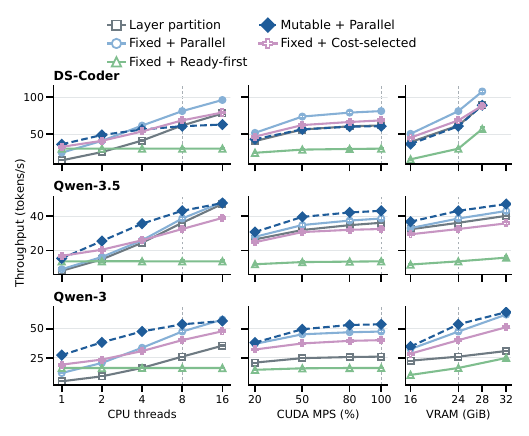}
  \caption{Measured throughput of five selected candidates under 10 A100 resource configurations.}

  \label{fig:hardware-tradeoffs}
\end{figure}

Before evaluating prediction, we ask whether one fixed policy could suffice.
Figure~\ref{fig:hardware-tradeoffs} crosses three MoE models with one-factor
A100 resource sweeps. The CPU sweep uses 1, 2, 4, 8, and 16 workers, while the
requested MPS sweep uses 20\%, 50\%, 80\%, and 100\%. The VRAM sweep uses caps
of 16, 24, and 28~GiB for DS-Coder and 16, 24, and 32~GiB for the two Qwen
models. Each sweep holds the other two controls at 8 workers, 100\% requested
MPS share, and 24~GiB of VRAM, as applicable. The figure plots the same
five-policy slice in every panel; Table~\ref{tab:winner-grid} reports the full
nine-candidate argmin at five representative resource configurations.

\begin{table}[t]
  \caption{Measured TPOT winner at representative
  A100 model--configuration points. Parentheses give speedup over layer
  partition, $\mathrm{TPOT}_{\mathrm{layer\ partition}} /
  \mathrm{TPOT}_{\mathrm{winner}}$. Each cell is tinted with its winner's policy
  color from Figures~\ref{fig:policy-landscape} and~\ref{fig:hardware-tradeoffs}.}
  \label{tab:winner-grid}
  \centering
  \scriptsize
  \setlength{\tabcolsep}{2pt}
  \begin{tabular}{@{}lccc@{}}
\toprule
Configuration & DS-Coder & Qwen-3.5 & Qwen-3 \\
\midrule
CPU threads = 1 & \cellcolor{pcMutableReadyFirst!60}\shortstack{Mutable + \\ Ready-first (2.62$\times$)} & \cellcolor{pcMutableCostSelected!60}\shortstack{Mutable + \\ Cost-selected (2.78$\times$)} & \cellcolor{pcMutableCostSelected!60}\shortstack{Mutable + \\ Cost-selected (6.05$\times$)} \\
CPU threads = 2 & \cellcolor{pcMutableParallel!60}\shortstack{Mutable + \\ Parallel (1.90$\times$)} & \cellcolor{pcMutableCostSelected!60}\shortstack{Mutable + \\ Cost-selected (1.74$\times$)} & \cellcolor{pcMutableParallel!60}\shortstack{Mutable + \\ Parallel (4.13$\times$)} \\
CPU threads = 16 & \cellcolor{pcFixedParallel!60}\shortstack{Fixed + \\ Parallel (1.23$\times$)} & \cellcolor{pcFixedParallel!60}\shortstack{Fixed + \\ Parallel (1.02$\times$)} & \cellcolor{pcFixedParallel!60}\shortstack{Fixed + \\ Parallel (1.63$\times$)} \\
MPS = 50\% & \cellcolor{pcFixedParallel!60}\shortstack{Fixed + \\ Parallel (1.31$\times$)} & \cellcolor{pcMutableParallel!60}\shortstack{Mutable + \\ Parallel (1.24$\times$)} & \cellcolor{pcMutableParallel!60}\shortstack{Mutable + \\ Parallel (1.99$\times$)} \\
VRAM = 16 GiB & \cellcolor{pcFixedParallel!60}\shortstack{Fixed + \\ Parallel (1.29$\times$)} & \cellcolor{pcMutableParallel!60}\shortstack{Mutable + \\ Parallel (1.14$\times$)} & \cellcolor{pcMutableParallel!60}\shortstack{Mutable + \\ Parallel (1.53$\times$)} \\
\bottomrule
\end{tabular}

\end{table}

The results expose a mismatch among models, configurations, and policies: four
candidates win across the 15 cells. With one CPU thread, DeepSeek picks
\emph{Mutable ready-first}, while Qwen models pick \emph{Mutable cost-selected}.
DeepSeek itself moves to \emph{Mutable Parallel} at two threads and \emph{Fixed Parallel} at 16.
Yet Qwen-3.5's 16-thread winner improves on \emph{layer partition} by only $1.02\times$.
Thus, neither the model nor the configuration alone determines the winner, and
policy selection must account for their interaction.

\textbf{Controlled single-model matrix.}
To isolate the effect of resource configuration from model differences, we
evaluate all nine candidates for DeepSeek-Coder-V2-Lite under four A100
configurations, comprising 540 runs. Figure~\ref{fig:policy-landscape} presents
the 405 runs from three of these configurations. The winning candidate changes
with CPU resources and cache capacity: mutable candidates win the two
low-worker configurations, whereas \emph{Fixed Parallel} wins both 16-worker
configurations. In the latter two cases, \emph{layer partition} is only
12.2--25.0\% slower, while \emph{wait-then-GPU} and \emph{ready-first} are
163--364\% slower. These results show that resource changes alone can reorder
the candidates even for a fixed model and physical GPU, and additional
staging or prediction mechanisms may not improve the end-to-end performance.

\subsection{Optimal Policy Prediction on A100}
\label{sec:qwen-a100-accuracy}

We next examine whether accurate policy selection requires measurements of
combined-operator schedules, rather than atomic-operator costs alone. We
compare COPE with an atomic-operator baseline across ten A100 resource
configurations for each of DS-Coder, Qwen-3.5, and Qwen-3. Each configuration
evaluates nine candidates using five prompts and five fresh-process trials.
All admitted runs remain within their VRAM caps and avoid swapping.

The atomic-operator baseline predicts the latency of each cache layer as
\[
\max(R[h],C[c])+\max(S[u],G[g])+A+F+M+\mathbf 1_{p>0}P.
\]
For layer partition, it estimates CPU layers as $C[k]+A+F+M$ and GPU layers
as $\max(R[k],G[k])+M$. This baseline represents concurrency through the two
maximum terms, without measuring the corresponding combined-operator execution
schedule. COPE and the baseline otherwise use the same common decode cost,
routing traces, replay states, and cost-selected inner assignments. The
comparison therefore isolates how the two methods compose the same underlying
costs into candidate-level TPOT predictions.

\begin{figure}[t]
  \centering
  \setlength{\belowcaptionskip}{-5pt}
  \includegraphics[width=\columnwidth]{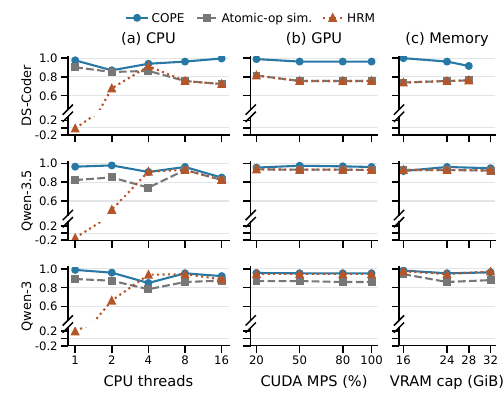}
  \caption{Mean Spearman rank correlation over the nine candidates on the A100 resource
  sweeps for DS-Coder, Qwen-3.5, and Qwen-3. Higher is better.}

  \label{fig:qwen-a100-ranking}
\end{figure}

We evaluate the resulting predictions at three levels. MAPE measures absolute
TPOT error, Figure~\ref{fig:qwen-a100-ranking} evaluates the predicted ordering
of all nine candidates, and Table~\ref{tab:qwen-a100-selection} reports whether
each method selects the candidate with the lowest measured TPOT. Together,
these metrics show whether schedule-level measurements improve not only latency
estimation, but also the policy-selection decision.

\begin{table}[t]
  \caption{Top-1 policy-selection accuracy across A100 resource sweeps for DS-Coder,
  Qwen-3.5, and Qwen-3. The measured lowest-TPOT policy among nine candidates is
  the ground truth for each prompt. Each $x/5$ reports the number of prompts for
  which the predicted and measured winners agree; dashes indicate infeasible
  configurations.}
  \label{tab:qwen-a100-selection}
  \centering
  \scriptsize
  \setlength{\tabcolsep}{1pt}
  \begin{tabular}{@{}ll cc ccc ccc ccc@{}}
\toprule
 & & & \multicolumn{4}{c}{threads} & \multicolumn{3}{c}{MPS} & \multicolumn{3}{c}{VRAM} \\
\multicolumn{2}{c}{(threads, MPS, VRAM)} & $(8,100,24)$ & \multicolumn{4}{c}{$(t,100,24)$} & \multicolumn{3}{c}{$(8,\mathrm{MPS},24)$} & \multicolumn{3}{c}{$(8,100,\mathrm{VRAM})$} \\
\cmidrule(lr){4-7}\cmidrule(lr){8-10}\cmidrule(lr){11-13}
 & & & 1 & 2 & 4 & 16 & 20 & 50 & 80 & 16 & 28 & 32 \\
\midrule
    \multirow{3}{*}{DS-Coder} & COPE & \textbf{5/5} & \textbf{1/5} & \textbf{5/5} & \textbf{4/5} & \textbf{5/5} & \textbf{5/5} & \textbf{5/5} & \textbf{5/5} & \textbf{5/5} & \textbf{5/5} & -- \\
     & Atomic-op sim. & 0/5 & 0/5 & \textbf{5/5} & 0/5 & 0/5 & 0/5 & 0/5 & 0/5 & 0/5 & 0/5 & -- \\
     & HRM & 0/5 & 0/5 & \textbf{5/5} & 0/5 & 0/5 & 0/5 & 0/5 & 0/5 & 0/5 & 0/5 & -- \\
    \multirow{3}{*}{Qwen-3.5} & COPE & \textbf{5/5} & \textbf{5/5} & \textbf{4/5} & \textbf{5/5} & \textbf{4/5} & \textbf{5/5} & \textbf{5/5} & \textbf{5/5} & \textbf{5/5} & -- & \textbf{5/5} \\
     & Atomic-op sim. & \textbf{5/5} & 0/5 & 1/5 & \textbf{5/5} & 0/5 & \textbf{5/5} & \textbf{5/5} & \textbf{5/5} & \textbf{5/5} & -- & \textbf{5/5} \\
     & HRM & \textbf{5/5} & 0/5 & 1/5 & \textbf{5/5} & 0/5 & \textbf{5/5} & \textbf{5/5} & \textbf{5/5} & \textbf{5/5} & -- & \textbf{5/5} \\
    \multirow{3}{*}{Qwen-3} & COPE & \textbf{5/5} & \textbf{4/5} & 2/5 & \textbf{5/5} & \textbf{2/5} & 4/5 & \textbf{5/5} & \textbf{5/5} & 4/5 & -- & \textbf{5/5} \\
     & Atomic-op sim. & \textbf{5/5} & 0/5 & \textbf{5/5} & \textbf{5/5} & \textbf{2/5} & \textbf{5/5} & \textbf{5/5} & \textbf{5/5} & \textbf{5/5} & -- & \textbf{5/5} \\
     & HRM & \textbf{5/5} & 0/5 & \textbf{5/5} & \textbf{5/5} & \textbf{2/5} & \textbf{5/5} & \textbf{5/5} & \textbf{5/5} & \textbf{5/5} & -- & \textbf{5/5} \\
\bottomrule
\end{tabular}

\end{table}

Across the A100 resource sweeps, schedule-level profiling generally predicts
the policy-performance landscape more accurately than composing
atomic-operator costs alone. For Qwen-3.5, COPE reduces MAPE from 15.12\% to
8.71\%, increases mean Spearman correlation from 0.884 to 0.944, and selects
the measured winner in 48 of the 50 cases, compared with
36 for the atomic-operator baseline.
Table~\ref{tab:qwen-a100-selection} extends the selection comparison to the
other two models. On DS-Coder, COPE selects 45 of the 50 measured winners,
whereas the baseline selects only 5. On Qwen-3, COPE selects 41 of the 50
measured winners, with a mean Spearman correlation of 0.949 and a MAPE of
29.52\%. Across the three models, these results show that combined-operator
schedule profiles can recover both measured winners and the broader candidate
ordering under changes in A100 resources.

\subsection{Optimal Policy Prediction on RTX 4090 and RTX 5090}
\label{sec:device-accuracy}

To determine whether the A100 results extend to a different physical platform,
we evaluate COPE on RTX~4090 and RTX~5090 under a fixed CPU allocation and full GPU
compute share. The study covers DeepSeek Coder, DeepSeek Chat, Qwen-3.5, and
Qwen3. For each model, we evaluate nine candidates using five prompts and five
fresh-process trials per candidate. The atomic-operator baseline uses the same
calibration primitives and routing-trace replays as COPE, isolating the
effect of schedule-level latency modeling.

\begin{figure}[t]
  \centering
  \setlength{\belowcaptionskip}{-5pt}
  \includegraphics[width=\columnwidth]{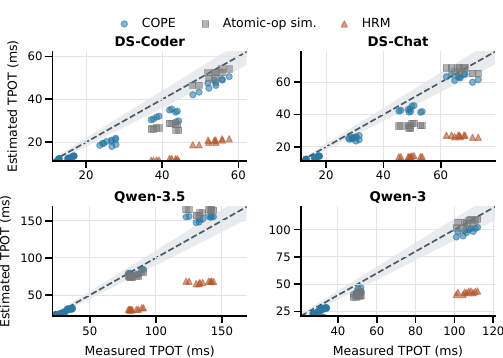}
  \caption{RTX~4090 predicted versus measured TPOT for four models, five
  prompts, and nine candidates per model. Each measured point averages five
  fresh processes. The dashed identity line and shaded $\pm$10\% band provide
  an absolute-error reference.}

  \label{fig:rtx4090-scatter}
\end{figure}

\begin{figure}[t]
  \centering
  \setlength{\belowcaptionskip}{-5pt}
  \includegraphics[width=\columnwidth]{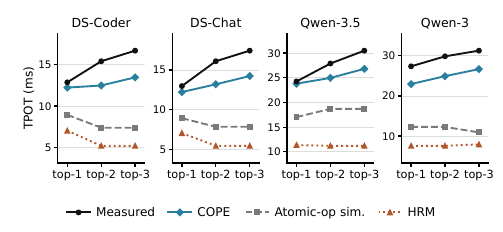}
  \caption{Measured and predicted TPOT of the three fastest measured candidates on the RTX~4090.}

  \label{fig:rtx4090-primitive-comparison}
\end{figure}

\begin{table}[t]
  \caption{RTX~4090 selection counts, rank correlation, and candidate error for
  all estimators. Top-1 compares predicted and measured TPOT argmins; Top-3 is
  three-policy set overlap; MAPE averages over all prompt--candidate pairs.}
  \label{tab:rtx4090}
  \centering
  \footnotesize
  \renewcommand{\arraystretch}{1.1}
  \setlength{\tabcolsep}{2.0pt}
  \begin{tabular}{@{}llrrrr@{}}
    \toprule
    Model & Estimator & Top-1 & Top-3 & $\rho$ & MAPE \\
    \midrule
    \multirow{3}{*}{DS-Coder} & COPE & \textbf{5/5} & \textbf{15/15} & \textbf{1.000} & \textbf{17.43\%} \\
      & Atomic-op sim. & 0/5 & 5/15 & 0.678 & 40.11\% \\
      & HRM & 0/5 & 5/15 & 0.678 & 67.38\% \\
    \multirow{3}{*}{DS-Chat} & COPE & \textbf{5/5} & \textbf{15/15} & \textbf{1.000} & \textbf{12.38\%} \\
      & Atomic-op sim. & 0/5 & 5/15 & 0.678 & 40.62\% \\
      & HRM & 0/5 & 5/15 & 0.678 & 67.77\% \\
    \multirow{3}{*}{Qwen-3.5} & COPE & \textbf{5/5} & \textbf{15/15} & \textbf{0.987} & \textbf{9.39\%} \\
      & Atomic-op sim. & 0/5 & 5/15 & 0.729 & 31.73\% \\
      & HRM & 0/5 & 5/15 & 0.678 & 61.53\% \\
    \multirow{3}{*}{Qwen-3} & COPE & \textbf{5/5} & \textbf{14/15} & \textbf{0.993} & \textbf{13.62\%} \\
      & Atomic-op sim. & 0/5 & 9/15 & 0.776 & 38.29\% \\
      & HRM & 0/5 & 9/15 & 0.776 & 68.52\% \\
    \bottomrule
  \end{tabular}
\end{table}

On the RTX~4090, Figure~\ref{fig:rtx4090-scatter} compares predicted and measured TPOT across
all candidates, while Figure~\ref{fig:rtx4090-primitive-comparison} focuses on
the three fastest measured candidates. Table~\ref{tab:rtx4090} summarizes
winner selection, candidate ordering, and absolute prediction error. Across
the 20 model--prompt cases, COPE selects every measured winner and recovers
59 of the 60 top-three slots. In comparison, the atomic-operator baseline
selects none of the measured winners and recovers 24 top-three slots. COPE
also reduces MAPE to 9.39--17.43\%, compared with 31.73--40.62\% for the
baseline. These results show that schedule-level profiling preserves policy
selection and ranking accuracy on a consumer GPU, rather than only within the
controlled A100 resource sweeps.

To complement the fixed-configuration RTX~4090 experiment, we evaluate COPE
across three VRAM budgets on the RTX~5090. We test DeepSeek Coder, DeepSeek
Chat, and Qwen3 with VRAM reservations of 16, 12, and 8~GiB, a fixed CPU
allocation, and full GPU compute share. Each configuration includes five
prompts and nine candidates. Qwen3 uses a Q8\_0 deployment with a separate
calibration.

\begin{figure}[t]
  \centering
  \setlength{\belowcaptionskip}{-5pt}
  \includegraphics[width=\columnwidth]{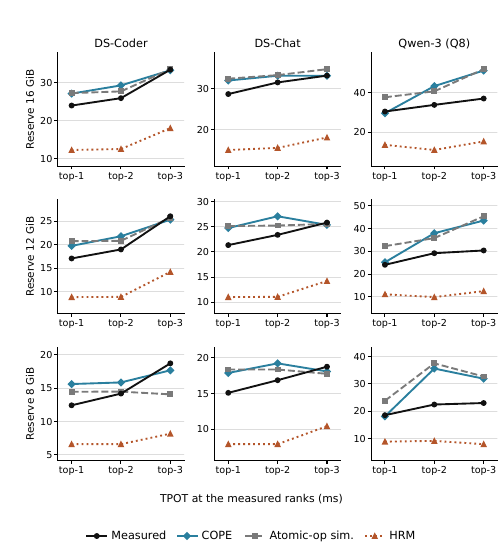}
  \caption{Measured and predicted TPOT of the three fastest measured candidates
  in each of nine model--reservation cells on the RTX~5090.}

  \label{fig:rtx5090-top3}
\end{figure}

\begin{table}[t]
  \caption{RTX~5090 prediction over three VRAM reservations per model.
  Denominators are 15 prompts and 45 top-three slots per model.}

  \label{tab:rtx5090}
  \centering
  \footnotesize
  \renewcommand{\arraystretch}{1.1}
  \setlength{\tabcolsep}{2pt}
  \begin{tabular}{@{}llrrrr@{}}
\toprule
Model & Estimator & Top-1 & Top-3 & $\rho$ & MAPE \\
\midrule
\multirow{3}{*}{DS-Coder} & COPE & \textbf{15/15} & \textbf{44/45} & \textbf{0.991} & \textbf{9.10\%} \\
 & Atomic-op sim. & 0/15 & 17/45 & 0.745 & 23.03\% \\
 & HRM & 0/15 & 17/45 & 0.745 & 53.81\% \\
\multirow{3}{*}{DS-Chat} & COPE & \textbf{12/15} & \textbf{45/45} & \textbf{0.979} & \textbf{7.95\%} \\
 & Atomic-op sim. & 0/15 & 15/45 & 0.719 & 24.13\% \\
 & HRM & 0/15 & 15/45 & 0.736 & 53.92\% \\
\multirow{3}{*}{Qwen-3 (Q8)} & COPE & \textbf{15/15} & 30/45 & 0.964 & \textbf{13.34\%} \\
 & Atomic-op sim. & \textbf{15/15} & \textbf{45/45} & \textbf{0.986} & 19.60\% \\
 & HRM & 0/15 & \textbf{45/45} & 0.962 & 47.21\% \\
\bottomrule
\end{tabular}

\end{table}

Figure~\ref{fig:rtx5090-top3} compares measured and predicted TPOT for the
three fastest candidates, while Table~\ref{tab:rtx5090} summarizes selection,
ranking, and absolute prediction error. COPE selects every measured winner
for Coder and Qwen3, and 12 of the 15 winners for Chat. It outperforms the
atomic-operator baseline on all three metrics for both DeepSeek models. For
Qwen3, both methods select every winner, but their other metrics diverge: the
atomic-operator baseline better preserves the candidate ordering, whereas
COPE provides lower absolute TPOT error.

The RTX~5090 results complement the fixed-configuration RTX~4090 study. COPE
shows consistently stronger selection and ranking on the RTX~4090, while the
RTX~5090 VRAM sweeps demonstrate that winner selection, candidate ordering,
and absolute TPOT error do not always improve together. Taken with the A100
results, these experiments show that prediction quality depends on both the
model and the available platform resources, motivating the separate reporting
of all three metrics.

\subsection{Engineering Effort and Calibration Cost}
\label{sec:prediction-cost}

\begin{figure}[t]
  \centering
  \setlength{\belowcaptionskip}{-5pt}
  \includegraphics[width=\columnwidth]{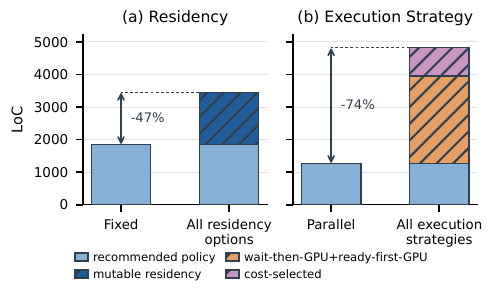}
  \caption{Policy-specific lines added to llama.cpp for the fixed-Parallel
  policy selected for Qwen3 on the RTX~4090, versus implementing every option
  along each Policy Matrix axis. Shared plumbing and diagnostics are excluded.}
  \label{fig:policy-implementation-cost}
\end{figure}

\begin{table}[t]
  \caption{RTX~4090 profiling time in minutes. Atomic-operator times are
  measured; combined-operator times are projected as 25\% of measured
  exhaustive-schedule time, assuming only replay-reachable states are profiled.}
  \label{tab:calibration-cost}
  \centering
  \footnotesize
  \renewcommand{\arraystretch}{1.1}
  \setlength{\tabcolsep}{4pt}
  \begin{tabular}{@{}lcccc@{}}
    \toprule
    Profile & DS-Coder & DS-Chat & Qwen-3.5 & Qwen-3 \\
    \midrule
    Atomic-operator profile  & 2.8 & 2.8 & 3.4 & 3.5 \\
    Combined ops  & 25.9 & 25.9 & 43.5 & 33.9 \\
    \bottomrule
  \end{tabular}
\end{table}

Accurate TPOT prediction can avoid implementing policies that will not be
deployed. For Qwen3 on the RTX~4090, COPE selects fixed Parallel.
Figure~\ref{fig:policy-implementation-cost} compares the policy-specific
lines added to llama.cpp for this choice with those needed to cover every
option in each Policy Matrix dimension. The choice would avoid 47\% of residency
code and 74\% of execution code. Across both dimensions, the selected policy requires 3,107 policy-specific lines of code,
whereas implementing all options requires 8,276, even when shared mechanisms are
counted only once.
Shared plumbing and prediction/diagnostic code are excluded from both totals.

Extending COPE's prediction path to a new backend is simpler than
implementing its offloading policies. Once the backend supports the atomic
operations needed for inference, it only needs to expose those operations
through profiling interfaces. VHM can then reuse its policy compositions
and trace replay to estimate TPOT without integrating each candidate as a
production offloader.

Profiling once on the target amortizes calibration across all candidate
policies. On the RTX~4090, atomic-operator profiles take 2.8--3.5 minutes
across four models (Table~\ref{tab:calibration-cost}).
Profiling combined-operator schedules expands the measurement space beyond
that of individual operators. Exhaustively covering this space takes
25.9--43.5 minutes.

\section{Related Work}

\textbf{Expert caches.}
MoE-Infinity, Mixtral-Offloading, ProMoE, DuoServe-MoE, and FreeToken study demand loading,
prediction, and overlap for sparse experts
\cite{moe-infinity,mixtral-offloading,promoe,duoserve-moe,freetoken}.
KTransformers uses fixed CPU-routed/GPU-shared placement in its full-accuracy
path \cite{ktransformers}, and SGLang KT integrates that execution approach
\cite{sglang-kt}. Fiddler selects CPU or GPU execution for misses using
per-expert input size, while FreeToken splits them by measured host and PCIe
bandwidth \cite{fiddler,freetoken}; HybriMoE and DALI also use workload or cost
information \cite{hybrimoe,dali}. COPE reuses these mechanism families. Its contribution
is to organize them as residency/execution candidates and predict which complete
template to integrate without using a production candidate's latency.

\textbf{Adjacent mechanisms.}
MoE-Lightning pages weights in batches, so its matrix placement captures a
mechanism rather than a batch-1 expert cache \cite{moe-lightning}. vLLM UVA
uses GPU remote reads from host memory \cite{vllm-uva}. vLLM Prefetch and SGLang
OffloaderV2 instead stage modules in execution order
\cite{vllm-prefetch,sglang-offloaderv2}. These latter mechanisms remain
deployment alternatives but do not share the candidate semantics measured by
the Policy Matrix.

\textbf{Other representation and routing choices.}
FineMoE moves weights at finer granularity, while ZipMoE applies lossless
compression~\cite{finemoe-offloading,zipmoe}. Both depart from our whole-expert
representation assumption without necessarily reducing accuracy. Approximate
routing, by contrast, changes the selected computation. COPE fixes both
whole-expert representation and exact routing to compare scheduling and
residency within one declared policy space.

\section{Discussion}

\textbf{TPOT rather than TTFT.}
COPE optimizes decode TPOT, not time to first token. Standard MoE execution
groups routed tokens by expert for batched or block-sparse matrix operations
\cite{megablocks}. Fiddler observes that longer prefills can make CPU expert
computation costlier than staging weights for GPU execution~\cite{fiddler}.
This shifts the CPU/GPU crossover; our batch-1 results do not predict prefill
performance.

\textbf{Batch $n$ versus batch 1.}
Batch-$n$ serving has a different CPU/GPU crossover. Multiple tokens can
share an expert~\cite{fiddler}; MoE-Lightning studies this regime with
layer-paged weights~\cite{moe-lightning}. It requires batch-size-specific
probes and may rank policies differently from single-user batch-1 decode.

\textbf{Unmodified computation versus accuracy-for-speed offloading.}
COPE's candidates preserve checkpoint routing, execute every selected expert,
and retain stored precision. Other systems predict sparsity or change routing
\cite{deja-vu,pregated-moe,moe-lookahead-gate}, change expert precision through
quantization
\cite{moe-apex,moqe,qmoe,mxmoe,mc-moe,automated-moe-quant,eac-moe,gemq}, or
skip, prune, substitute, or merge experts or neurons
\cite{smoe,not-all-experts-equal,moe-i2,adapmoe,sub-moe,dern,
diversifying-expert-knowledge,shapley-moe}. Their accuracy--performance
trade-offs are not directly comparable with TPOT for an unchanged checkpoint;
the three invariants define our exactness and quality requirements.

\textbf{Disk versus host-memory offloading.}
COPE assumes exact routing, whole experts, and a model fitting in host RAM.
FlexGen instead spans GPU, CPU, and disk, staging disk-resident tensors through
DRAM~\cite{flexgen}. Disk requires another set of templates and probes; on a
shared-memory AI SoC, it becomes a capacity tier analogous to host RAM below
discrete-GPU HBM.

\section{Conclusion}
MoE offloading performance depends jointly on expert residency and execution
strategy, making policy selection specific to each model--platform
configuration. COPE structures this design space with a Policy Matrix and
evaluates candidates before integration using a Virtual Heterogeneous Machine.
By replacing implementation-intensive trial and error with systematic
pre-integration analysis, COPE lowers the engineering barrier to deploying MoE
models on heterogeneous, resource-constrained hardware.

\bibliographystyle{ACM-Reference-Format}
\bibliography{eurosys27-hot-cache}

@article{switch-transformers,
  title = {{Switch Transformers: Scaling to Trillion Parameter Models with Simple and Efficient Sparsity}},
  author = {William Fedus and Barret Zoph and Noam Shazeer},
  journal = {Journal of Machine Learning Research},
  volume = {23},
  number = {120},
  pages = {1--39},
  year = {2022},
  url = {https://jmlr.org/papers/v23/21-0998.html}
}

@misc{mixtral,
  title = {{Mixtral of Experts}},
  author = {Albert Q. Jiang and others},
  year = {2024},
  eprint = {2401.04088},
  archivePrefix = {arXiv},
  primaryClass = {cs.LG}
}

@misc{deepseek-v2,
  title = {{DeepSeek-V2: A Strong, Economical, and Efficient Mixture-of-Experts Language Model}},
  author = {{DeepSeek-AI}},
  year = {2024},
  eprint = {2405.04434},
  archivePrefix = {arXiv},
  primaryClass = {cs.CL}
}

@misc{deepseek-coder-v2,
  title = {{DeepSeek-Coder-V2: Breaking the Barrier of Closed-Source Models in Code Intelligence}},
  author = {{DeepSeek-AI}},
  year = {2024},
  eprint = {2406.11931},
  archivePrefix = {arXiv},
  primaryClass = {cs.SE}
}

@misc{qwen3,
  title = {{Qwen3 Technical Report}},
  author = {An Yang and others},
  year = {2025},
  eprint = {2505.09388},
  archivePrefix = {arXiv},
  primaryClass = {cs.CL}
}

@misc{qwen35,
  title = {{Qwen3.5: Towards Native Multimodal Agents}},
  author = {{Qwen Team}},
  month = feb,
  year = {2026},
  url = {https://qwen.ai/blog?id=qwen3.5}
}

@misc{deepseek-coder-v2-lite-card,
  title = {{DeepSeek-Coder-V2-Lite-Instruct Model Card}},
  author = {{DeepSeek-AI}},
  year = {2024},
  url = {https://huggingface.co/deepseek-ai/DeepSeek-Coder-V2-Lite-Instruct},
  note = {Official checkpoint documentation, accessed 25 September 2026}
}

@misc{deepseek-v2-lite-chat-card,
  title = {{DeepSeek-V2-Lite-Chat Model Card}},
  author = {{DeepSeek-AI}},
  year = {2024},
  url = {https://huggingface.co/deepseek-ai/DeepSeek-V2-Lite-Chat},
  note = {Official checkpoint documentation, accessed 25 September 2026}
}

@misc{qwen3-30b-a3b-card,
  title = {{Qwen3-30B-A3B Model Card}},
  author = {{Qwen Team}},
  year = {2025},
  url = {https://huggingface.co/Qwen/Qwen3-30B-A3B},
  note = {Official checkpoint documentation, accessed 25 September 2026}
}

@misc{qwen35-35b-a3b-card,
  title = {{Qwen3.5-35B-A3B Model Card}},
  author = {{Qwen Team}},
  year = {2026},
  url = {https://huggingface.co/Qwen/Qwen3.5-35B-A3B},
  note = {Official checkpoint documentation, accessed 25 September 2026}
}

@misc{moe-infinity,
  title = {{MoE-Infinity: Efficient MoE Inference on Personal Machines with Sparsity-Aware Expert Cache}},
  author = {Leyang Xue and Yao Fu and Zhan Lu and Luo Mai and Mahesh Marina},
  year = {2024},
  eprint = {2401.14361},
  archivePrefix = {arXiv},
  primaryClass = {cs.LG}
}

@inproceedings{finemoe-offloading,
  title = {{Taming Latency-Memory Trade-Off in MoE-Based LLM Serving via Fine-Grained Expert Offloading}},
  author = {Hanfei Yu and Xingqi Cui and Hong Zhang and Hao Wang and Hao Wang},
  booktitle = {Proceedings of the 21st European Conference on Computer Systems},
  publisher = {Association for Computing Machinery},
  address = {New York, NY, USA},
  pages = {176--191},
  year = {2026},
  doi = {10.1145/3767295.3769319},
  url = {https://doi.org/10.1145/3767295.3769319}
}

@misc{mixtral-offloading,
  title = {{Fast Inference of Mixture-of-Experts Language Models with Offloading}},
  author = {Artyom Eliseev and Denis Mazur},
  year = {2023},
  eprint = {2312.17238},
  archivePrefix = {arXiv},
  primaryClass = {cs.LG}
}

@inproceedings{fiddler,
  title = {{Fiddler: CPU-GPU Orchestration for Fast Inference of Mixture-of-Experts Models}},
  author = {Keisuke Kamahori and Tian Tang and Yile Gu and Kan Zhu and Baris Kasikci},
  booktitle = {The Thirteenth International Conference on Learning Representations},
  year = {2025},
  url = {https://proceedings.iclr.cc/paper_files/paper/2025/hash/8cd1ce03ea58b3d7dfd809e4d42f08ea-Abstract-Conference.html}
}

@misc{promoe,
  title = {{ProMoE: Fast MoE-based LLM Serving using Proactive Caching}},
  author = {Xiaoniu Song and Zihang Zhong and Rong Chen and Haibo Chen},
  year = {2024},
  eprint = {2410.22134},
  archivePrefix = {arXiv},
  primaryClass = {cs.DC}
}

@misc{moe-lightning,
  title = {{MoE-Lightning: High-Throughput MoE Inference on Memory-constrained GPUs}},
  author = {Shiyi Cao and Shu Liu and Tyler Griggs and Peter Schafhalter and Xiaoxuan Liu and Ying Sheng and Joseph E. Gonzalez and Matei Zaharia and Ion Stoica},
  year = {2024},
  eprint = {2411.11217},
  archivePrefix = {arXiv},
  primaryClass = {cs.DC}
}

@misc{freetoken,
  title = {{FreeToken: Efficient Edge-Native MoE Serving with Bandwidth-Adaptive Execution}},
  author = {Shuo Yang and Xiaoze Fan and Melissa Pan and Haocheng Xi and Zhe Wang and Shanlin Sun and Kurt Keutzer and Song Han and Matei Zaharia and Chenfeng Xu and Ion Stoica},
  year = {2026},
  eprint = {2608.16157},
  archivePrefix = {arXiv},
  primaryClass = {cs.DC}
}

@inproceedings{deja-vu,
  title = {{Deja Vu: Contextual Sparsity for Efficient LLMs at Inference Time}},
  author = {Zichang Liu and Jue Wang and Tri Dao and Tianyi Zhou and Binhang Yuan and Zhao Song and Anshumali Shrivastava and Ce Zhang and Yuandong Tian and Christopher R{\'e} and Beidi Chen},
  booktitle = {Proceedings of the 40th International Conference on Machine Learning},
  series = {Proceedings of Machine Learning Research},
  volume = {202},
  pages = {22137--22176},
  publisher = {PMLR},
  year = {2023},
  url = {https://proceedings.mlr.press/v202/liu23am.html}
}

@inproceedings{hybrimoe,
  title = {{HybriMoE: Hybrid CPU-GPU Scheduling and Cache Management for Efficient MoE Inference}},
  author = {Shuzhang Zhong and Yanfan Sun and Ling Liang and Runsheng Wang and Ru Huang and Meng Li},
  booktitle = {Proceedings of the 62nd ACM/IEEE Design Automation Conference (DAC)},
  publisher = {IEEE},
  address = {San Francisco, CA, USA},
  pages = {1--7},
  numpages = {7},
  year = {2025},
  doi = {10.1109/DAC63849.2025.11133274},
  url = {https://doi.org/10.1109/DAC63849.2025.11133274}
}

@misc{dali,
  title = {{DALI: A Workload-Aware Offloading Framework for Efficient MoE Inference on Local PCs}},
  author = {Zeyu Zhu and Gang Li and Peisong Wang and Zitao Mo and Minnan Pei and Zhuoran Song and Xiaoyao Liang and Jian Cheng},
  year = {2026},
  eprint = {2602.03495},
  archivePrefix = {arXiv},
  primaryClass = {cs.DC}
}

@misc{duoserve-moe,
  title = {{DuoServe-MoE: Dual-Phase Expert Prefetch and Caching for LLM Inference QoS Assurance}},
  author = {Yuning Zhang and Grant Pinkert and Nan Yang and Yanli Li and Dong Yuan},
  year = {2025},
  eprint = {2509.07379},
  archivePrefix = {arXiv},
  primaryClass = {cs.DC},
  note = {arXiv:2509.07379v2, revised 9 April 2026}
}

@misc{llama-cpp,
  title = {{llama.cpp}},
  author = {{llama.cpp Contributors}},
  year = {2026},
  howpublished = {Source repository},
  url = {https://github.com/ggml-org/llama.cpp},
  note = {Accessed 2 September 2026}
}

@misc{ggml,
  title = {{ggml: Tensor Library for Machine Learning}},
  author = {{ggml Contributors}},
  year = {2026},
  howpublished = {Source repository},
  url = {https://github.com/ggml-org/ggml},
  note = {Accessed 25 September 2026}
}

@misc{cuda-async,
  title = {{CUDA Programming Guide: Asynchronous Execution}},
  author = {{NVIDIA}},
  year = {2026},
  url = {https://docs.nvidia.com/cuda/cuda-programming-guide/02-basics/asynchronous-execution.html},
  note = {Accessed 25 September 2026}
}

@misc{cuda-memory,
  title = {{CUDA Programming Guide: Unified and System Memory}},
  author = {{NVIDIA}},
  year = {2026},
  url = {https://docs.nvidia.com/cuda/cuda-programming-guide/02-basics/understanding-memory.html},
  note = {Accessed 25 September 2026}
}

@misc{cuda-mps,
  title = {{Multi-Process Service: Environment Variables}},
  author = {{NVIDIA}},
  year = {2026},
  url = {https://docs.nvidia.com/deploy/mps/appendix-environment-variables.html},
  note = {Accessed 25 September 2026}
}

@misc{tensorrt-llm,
  title = {{TensorRT-LLM Documentation}},
  author = {{NVIDIA}},
  year = {2026},
  url = {https://docs.nvidia.com/tensorrt-llm/},
  note = {Accessed 25 September 2026}
}

@inproceedings{vllm-sosp,
  title = {{Efficient Memory Management for Large Language Model Serving with PagedAttention}},
  author = {Woosuk Kwon and Zhuohan Li and Siyuan Zhuang and Ying Sheng and Lianmin Zheng and Cody Hao Yu and Joseph E. Gonzalez and Hao Zhang and Ion Stoica},
  booktitle = {Proceedings of the 29th Symposium on Operating Systems Principles},
  pages = {611--626},
  publisher = {Association for Computing Machinery},
  address = {New York, NY, USA},
  year = {2023},
  doi = {10.1145/3600006.3613165}
}

@inproceedings{sglang-paper,
  title = {{SGLang: Efficient Execution of Structured Language Model Programs}},
  author = {Lianmin Zheng and Liangsheng Yin and Zhiqiang Xie and Chuyue Sun and Jeff Huang and Cody Hao Yu and Shiyi Cao and Christos Kozyrakis and Ion Stoica and Joseph E. Gonzalez and Clark Barrett and Ying Sheng},
  booktitle = {Advances in Neural Information Processing Systems},
  volume = {37},
  publisher = {Curran Associates, Inc.},
  address = {Red Hook, NY, USA},
  year = {2024},
  doi = {10.52202/079017-2000},
  url = {https://proceedings.neurips.cc/paper_files/paper/2024/hash/724be4472168f31ba1c9ac630f15dec8-Abstract-Conference.html}
}

@inproceedings{pytorch,
  title = {{PyTorch: An Imperative Style, High-Performance Deep Learning Library}},
  author = {Adam Paszke and Sam Gross and Francisco Massa and Adam Lerer and James Bradbury and Gregory Chanan and Trevor Killeen and Zeming Lin and Natalia Gimelshein and Luca Antiga and others},
  booktitle = {Advances in Neural Information Processing Systems},
  volume = {32},
  pages = {8024--8035},
  publisher = {Curran Associates, Inc.},
  address = {Red Hook, NY, USA},
  year = {2019},
  url = {https://proceedings.neurips.cc/paper/2019/hash/bdbca288fee7f92f2bfa9f7012727740-Abstract.html}
}

@article{megablocks,
  title = {{MegaBlocks: Efficient Sparse Training with Mixture-of-Experts}},
  author = {Trevor Gale and Deepak Narayanan and Cliff Young and Matei Zaharia},
  journal = {Proceedings of Machine Learning and Systems},
  volume = {5},
  pages = {288--304},
  year = {2023},
  url = {https://proceedings.mlsys.org/paper_files/paper/2023/hash/5a54f79333768effe7e8927bcccffe40-Abstract-mlsys2023.html}
}

@inproceedings{flexgen,
  title = {{FlexGen: High-Throughput Generative Inference of Large Language Models with a Single GPU}},
  author = {Ying Sheng and Lianmin Zheng and Binhang Yuan and Zhuohan Li and Max Ryabinin and Beidi Chen and Percy Liang and Christopher R{\'e} and Ion Stoica and Ce Zhang},
  booktitle = {Proceedings of the 40th International Conference on Machine Learning},
  series = {Proceedings of Machine Learning Research},
  volume = {202},
  pages = {31094--31116},
  year = {2023},
  publisher = {PMLR},
  address = {Honolulu, HI, USA},
  url = {https://proceedings.mlr.press/v202/sheng23a.html}
}

@article{vidur,
  title = {{VIDUR: A Large-Scale Simulation Framework for LLM Inference}},
  author = {Amey Agrawal and Nitin Kedia and Jayashree Mohan and Ashish Panwar and Nipun Kwatra and Bhargav S. Gulavani and Ramachandran Ramjee and Alexey Tumanov},
  journal = {Proceedings of Machine Learning and Systems},
  volume = {6},
  numpages = {16},
  year = {2024},
  url = {https://proceedings.mlsys.org/paper_files/paper/2024/hash/b74a8de47d2b3c928360e0a011f48351-Abstract-Conference.html}
}

@inproceedings{llmservingsim2,
  title = {{LLMServingSim 2.0: A Unified Simulator for Heterogeneous and Disaggregated LLM Serving Infrastructure}},
  author = {Jaehong Cho and Hyunmin Choi and Guseul Heo and Jongse Park},
  booktitle = {2026 IEEE International Symposium on Performance Analysis of Systems and Software (ISPASS)},
  pages = {1--14},
  year = {2026},
  publisher = {IEEE},
  address = {Piscataway, NJ, USA},
  doi = {10.1109/ISPASS69572.2026.00012}
}

@inproceedings{frontier,
  title = {{Frontier: Simulating the Next Generation of LLM Inference Systems}},
  author = {Yicheng Feng and Xin Tan and Kin Hang Sew and Yimin Jiang and Yibo Zhu and Hong Xu},
  booktitle = {Proceedings of the 4th Workshop on Practical Adoption Challenges of ML for Systems (PACMI)},
  pages = {25--30},
  year = {2025},
  publisher = {Association for Computing Machinery},
  address = {New York, NY, USA},
  doi = {10.1145/3766882.3767173}
}

@inproceedings{simai,
  title = {{SimAI}: Unifying Architecture Design and Performance Tuning for {Large-Scale} Large Language Model Training with Scalability and Precision},
  author = {Xizheng Wang and Qingxu Li and Yichi Xu and Gang Lu and Dan Li and Li Chen and Heyang Zhou and Linkang Zheng and Sen Zhang and Yikai Zhu and Yang Liu and Pengcheng Zhang and Kun Qian and Kunling He and Jiaqi Gao and Ennan Zhai and Dennis Cai and Binzhang Fu},
  booktitle = {22nd USENIX Symposium on Networked Systems Design and Implementation (NSDI 25)},
  pages = {541--558},
  year = {2025},
  publisher = {USENIX Association},
  address = {Philadelphia, PA, USA},
  url = {https://www.usenix.org/conference/nsdi25/presentation/wang-xizheng-simai}
}

@inproceedings{astrasim2,
  title = {{ASTRA-sim2.0: Modeling Hierarchical Networks and Disaggregated Systems for Large-model Training at Scale}},
  author = {William Won and Taekyung Heo and Saeed Rashidi and Srinivas Sridharan and Sudarshan Srinivasan and Tushar Krishna},
  booktitle = {2023 IEEE International Symposium on Performance Analysis of Systems and Software (ISPASS)},
  pages = {283--294},
  year = {2023},
  publisher = {IEEE},
  address = {Piscataway, NJ, USA},
  doi = {10.1109/ISPASS57527.2023.00035}
}

@inproceedings{ktransformers,
  title = {{KTransformers: Unleashing the Full Potential of CPU/GPU Hybrid Inference for MoE Models}},
  author = {Hongtao Chen and Weiyu Xie and Boxin Zhang and Jingqi Tang and Jiahao Wang and Jianwei Dong and Shaoyuan Chen and Ziwei Yuan and Chen Lin and Chengyu Qiu and Yuening Zhu and Qingliang Ou and Jiaqi Liao and Xianglin Chen and Zhiyuan Ai and Yongwei Wu and Mingxing Zhang},
  booktitle = {Proceedings of the ACM SIGOPS 31st Symposium on Operating Systems Principles},
  pages = {1014--1029},
  year = {2025},
  publisher = {Association for Computing Machinery},
  address = {New York, NY, USA},
  doi = {10.1145/3731569.3764843}
}

@misc{sglang-kt,
  title = {{KT Expert Parallelism Wrapper for MoE Layers}},
  author = {{SGLang Contributors}},
  year = {2026},
  howpublished = {SGLang source repository},
  url = {https://github.com/sgl-project/sglang/blob/d97b796c168bdc764a3c70be388bd199469ec9ba/python/sglang/srt/layers/moe/kt_ep_wrapper.py},
  note = {Revision d97b796c, accessed 2026-08-17}
}

@misc{sglang-offloaderv2,
  title = {{SGLang OffloaderV2}},
  author = {{SGLang Contributors}},
  year = {2026},
  howpublished = {SGLang source repository},
  url = {https://github.com/sgl-project/sglang/blob/d97b796c168bdc764a3c70be388bd199469ec9ba/python/sglang/srt/utils/offloader.py},
  note = {Revision d97b796c, accessed 2026-08-17}
}

@misc{vllm-prefetch,
  title = {{vLLM Prefetch Offloader}},
  author = {{vLLM Contributors}},
  year = {2026},
  howpublished = {vLLM source repository},
  url = {https://github.com/vllm-project/vllm/blob/1d3a8b9e220f1edc77c190c3370cf0f76dfdd2fd/vllm/model_executor/offloader/prefetch.py},
  note = {Revision 1d3a8b9e, accessed 2026-08-17}
}

@misc{vllm-uva,
  title = {{vLLM UVA Offloader}},
  author = {{vLLM Contributors}},
  year = {2026},
  howpublished = {vLLM documentation and source repository},
  url = {https://docs.vllm.ai/en/latest/api/vllm/model_executor/offloader/uva/},
  note = {Accessed 2026-08-17}
}

@inproceedings{zipmoe,
  title={ZipMoE: Efficient On-Device MoE Serving via Lossless Compression and Cache-Affinity Scheduling},
  author={Yuchen Yang and Yaru Zhao and Pu Yang and Shaowei Wang and Zhi-Hua Zhou},
  booktitle={Proceedings of the 43rd International Conference on Machine Learning (ICML)},
  year={2026},
  url={https://arxiv.org/abs/2601.21198}
}

@inproceedings{pregated-moe,
  title = {{Pre-gated MoE: An Algorithm-System Co-Design for Fast and Scalable Mixture-of-Expert Inference}},
  author = {Ranggi Hwang and Jianyu Wei and Shijie Cao and Changho Hwang and Xiaohu Tang and Ting Cao and Mao Yang},
  booktitle = {2024 ACM/IEEE 51st Annual International Symposium on Computer Architecture (ISCA)},
  pages = {1018--1031},
  publisher = {IEEE},
  year = {2024},
  doi = {10.1109/ISCA59077.2024.00078}
}

@inproceedings{moe-apex,
  title = {{MoE-APEX: An Efficient MoE Inference System with Adaptive Precision Expert Offloading}},
  author = {Peng Tang and Jiacheng Liu and Xiaofeng Hou and Yifei Pu and Jing Wang and Pheng-Ann Heng and Chao Li and Minyi Guo},
  booktitle = {Proceedings of the 31st ACM International Conference on Architectural Support for Programming Languages and Operating Systems},
  volume = {2},
  pages = {1185--1200},
  publisher = {Association for Computing Machinery},
  year = {2026},
  doi = {10.1145/3779212.3790187}
}

@article{moe-lookahead-gate,
  title = {{Accelerating Mixture-of-Experts Language Model Inference via Plug-and-Play Lookahead Gate on a Single GPU}},
  author = {Jie Ou and Yueming Chen and Buyao Xiong and Zhaokun Wang and Wenhong Tian},
  journal = {Computer Standards \& Interfaces},
  volume = {94},
  pages = {103996},
  year = {2025},
  doi = {10.1016/j.csi.2025.103996}
}

@inproceedings{moqe,
  title = {{Mixture of Quantized Experts (MoQE): Complementary Effect of Low-bit Quantization and Robustness}},
  author = {Young Jin Kim and Raffy Fahim and Hany Hassan Awadalla},
  booktitle = {Efficient Natural Language and Speech Processing Workshop at NeurIPS},
  year = {2023},
  url = {https://neurips2023-enlsp.github.io/papers/paper_81.pdf}
}

@article{qmoe,
  title = {{QMoE: Sub-1-Bit Compression of Trillion Parameter Models}},
  author = {Elias Frantar and Dan Alistarh},
  journal = {Proceedings of Machine Learning and Systems},
  volume = {6},
  year = {2024},
  url = {https://proceedings.mlsys.org/paper_files/paper/2024/hash/c74b624843218d9b6713fcf299d6d5e4-Abstract-Conference.html}
}

@inproceedings{mxmoe,
  title = {{MxMoE: Mixed-precision Quantization for MoE with Accuracy and Performance Co-Design}},
  author = {Haojie Duanmu and Xiuhong Li and Zhihang Yuan and Size Zheng and Jiangfei Duan and Xingcheng Zhang and Dahua Lin},
  booktitle = {Proceedings of the 42nd International Conference on Machine Learning},
  series = {Proceedings of Machine Learning Research},
  volume = {267},
  pages = {14793--14806},
  publisher = {PMLR},
  year = {2025},
  url = {https://proceedings.mlr.press/v267/duanmu25a.html}
}

@inproceedings{mc-moe,
  title = {{Mixture Compressor for Mixture-of-Experts LLMs Gains More}},
  author = {Wei Huang and Yue Liao and Jianhui Liu and Ruifei He and Haoru Tan and Shiming Zhang and Hongsheng Li and Si Liu and Xiaojuan Qi},
  booktitle = {The Thirteenth International Conference on Learning Representations},
  year = {2025},
  url = {https://openreview.net/forum?id=hheFYjOsWO}
}

@inproceedings{automated-moe-quant,
  title = {{Automated Fine-Grained Mixture-of-Experts Quantization}},
  author = {Zhanhao Xie and Yuexiao Ma and Xiawu Zheng and Fei Chao and Wanchen Sui and Yong Li and Shen Li and Rongrong Ji},
  booktitle = {Findings of the Association for Computational Linguistics: ACL 2025},
  pages = {27024--27037},
  publisher = {Association for Computational Linguistics},
  year = {2025},
  doi = {10.18653/v1/2025.findings-acl.1386}
}

@inproceedings{eac-moe,
  title = {{EAC-MoE: Expert-Selection Aware Compressor for Mixture-of-Experts Large Language Models}},
  author = {Yuanteng Chen and Yuantian Shao and Peisong Wang and Jian Cheng},
  booktitle = {Proceedings of the 63rd Annual Meeting of the Association for Computational Linguistics},
  pages = {12942--12963},
  publisher = {Association for Computational Linguistics},
  year = {2025},
  doi = {10.18653/v1/2025.acl-long.633}
}

@inproceedings{gemq,
  title = {{GEMQ: Global Expert-Level Mixed-Precision Quantization for MoE LLMs}},
  author = {Jianing Deng and Song Wang and Dongwei Wang and Zijie Liu and Tianlong Chen and Huanrui Yang and Jingtong Hu},
  booktitle = {Proceedings of the 43rd International Conference on Machine Learning},
  year = {2026},
  url = {https://arxiv.org/abs/2605.23078}
}

@inproceedings{smoe,
  title = {{SMoE: An Algorithm-System Co-Design for Pushing MoE to the Edge via Expert Substitution}},
  author = {Guoying Zhu and Meng Li and Haipeng Dai and Xuechen Liu and Weijun Wang and Keran Li and Jun Xiao and Ligeng Chen and Wei Wang},
  booktitle = {2026 ACM/IEEE 53rd Annual International Symposium on Computer Architecture (ISCA)},
  pages = {1302--1318},
  publisher = {IEEE},
  year = {2026},
  doi = {10.1109/ISCA66397.2026.00099}
}

@inproceedings{not-all-experts-equal,
  title = {{Not All Experts are Equal: Efficient Expert Pruning and Skipping for Mixture-of-Experts Large Language Models}},
  author = {Xudong Lu and Qi Liu and Yuhui Xu and Aojun Zhou and Siyuan Huang and Bo Zhang and Junchi Yan and Hongsheng Li},
  booktitle = {Proceedings of the 62nd Annual Meeting of the Association for Computational Linguistics},
  pages = {6159--6172},
  publisher = {Association for Computational Linguistics},
  year = {2024},
  doi = {10.18653/v1/2024.acl-long.334}
}

@inproceedings{moe-i2,
  title = {{MoE-I2: Compressing Mixture of Experts Models through Inter-Expert Pruning and Intra-Expert Low-Rank Decomposition}},
  author = {Cheng Yang and Yang Sui and Jinqi Xiao and Lingyi Huang and Yu Gong and Yuanlin Duan and Wenqi Jia and Miao Yin and Yu Cheng and Bo Yuan},
  booktitle = {Findings of the Association for Computational Linguistics: EMNLP 2024},
  pages = {10456--10466},
  publisher = {Association for Computational Linguistics},
  year = {2024},
  doi = {10.18653/v1/2024.findings-emnlp.612}
}

@inproceedings{adapmoe,
  title = {{AdapMoE: Adaptive Sensitivity-based Expert Gating and Management for Efficient MoE Inference}},
  author = {Shuzhang Zhong and Ling Liang and Yuan Wang and Runsheng Wang and Ru Huang and Meng Li},
  booktitle = {Proceedings of the 43rd IEEE/ACM International Conference on Computer-Aided Design},
  articleno = {51},
  numpages = {9},
  publisher = {Association for Computing Machinery},
  year = {2024},
  doi = {10.1145/3676536.3676741}
}

@article{sub-moe,
  title = {{Sub-MoE: Efficient Mixture-of-Expert LLMs Compression via Subspace Expert Merging}},
  author = {Lujun Li and Qiyuan Zhu and Jiacheng Wang and Xiaoyu Qin and Wei Li and Hao Gu and Sirui Han and Yike Guo},
  journal = {Proceedings of the AAAI Conference on Artificial Intelligence},
  volume = {40},
  number = {27},
  pages = {22994--23002},
  year = {2026},
  doi = {10.1609/aaai.v40i27.39464}
}

@inproceedings{dern,
  title = {{Dropping Experts, Recombining Neurons: Retraining-Free Pruning for Sparse Mixture-of-Experts LLMs}},
  author = {Yixiao Zhou and Ziyu Zhao and Dongzhou Cheng and Zhiliang Wu and Jie Gui and Yi Yang and Fei Wu and Yu Cheng and Hehe Fan},
  booktitle = {Findings of the Association for Computational Linguistics: EMNLP 2025},
  pages = {15169--15186},
  publisher = {Association for Computational Linguistics},
  year = {2025},
  doi = {10.18653/v1/2025.findings-emnlp.820}
}

@inproceedings{diversifying-expert-knowledge,
  title = {{Diversifying the Expert Knowledge for Task-Agnostic Pruning in Sparse Mixture-of-Experts}},
  author = {Zeliang Zhang and Xiaodong Liu and Hao Cheng and Chenliang Xu and Jianfeng Gao},
  booktitle = {Findings of the Association for Computational Linguistics: ACL 2025},
  pages = {86--102},
  publisher = {Association for Computational Linguistics},
  year = {2025},
  doi = {10.18653/v1/2025.findings-acl.4}
}

@inproceedings{shapley-moe,
  title = {{Discovering Important Experts for Mixture-of-Experts Models Pruning Through a Theoretical Perspective}},
  author = {Weizhong Huang and Yuxin Zhang and Xiawu Zheng and Fei Chao and Rongrong Ji and Liujuan Cao},
  booktitle = {Advances in Neural Information Processing Systems},
  volume = {38},
  year = {2025},
  doi = {10.52202/085713-4538}
}

@inproceedings{math,
  title = {{Measuring Mathematical Problem Solving With the MATH Dataset}},
  author = {Dan Hendrycks and Collin Burns and Saurav Kadavath and Akul Arora and Steven Basart and Eric Tang and Dawn Song and Jacob Steinhardt},
  booktitle = {Advances in Neural Information Processing Systems: Datasets and Benchmarks Track},
  year = {2021}
}

@inproceedings{swe-bench,
  title = {{SWE-bench: Can Language Models Resolve Real-World GitHub Issues?}},
  author = {Carlos E. Jimenez and John Yang and Alexander Wettig and Shunyu Yao and Kexin Pei and Ofir Press and Karthik Narasimhan},
  booktitle = {International Conference on Learning Representations},
  year = {2024}
}

@misc{sharegpt,
  title = {{ShareGPT52K: Realistic Human-ChatGPT Conversations}},
  author = {{Ryoko AI}},
  year = {2023},
  howpublished = {\url{https://huggingface.co/datasets/RyokoAI/ShareGPT52K}}
}

@misc{gsm8k,
  title = {{Training Verifiers to Solve Math Word Problems}},
  author = {Karl Cobbe and Vineet Kosaraju and Mohammad Bavarian and Mark Chen and Heewoo Jun and Lukasz Kaiser and Matthias Plappert and Jerry Tworek and Jacob Hilton and Reiichiro Nakano and Christopher Hesse and John Schulman},
  year = {2021}, eprint = {2110.14168}, archivePrefix = {arXiv}, primaryClass = {cs.LG}
}

@misc{magicoder-evol-instruct,
  title = {{Magicoder-Evol-Instruct-110K}},
  author = {{Intelligent Software Engineering Lab, UIUC}},
  year = {2023},
  howpublished = {\url{https://huggingface.co/datasets/ise-uiuc/Magicoder-Evol-Instruct-110K}}
}

@misc{ultrachat-200k,
  title = {{UltraChat 200k}},
  author = {{Hugging Face H4}},
  year = {2023},
  howpublished = {\url{https://huggingface.co/datasets/HuggingFaceH4/ultrachat_200k}}
}

\end{document}